%% file: main.tex
\RequirePackage{pdfmanagement-testphase}
\DeclareDocumentMetadata{uncompress, pdfversion=1.7, pdfstandard=A-2b}

\documentclass[a4paper, 10pt, conference]{ieeeconf}      
\usepackage[colorlinks=true, linkcolor=blue, citecolor=blue, urlcolor=blue, backref=true]{hyperref}

\usepackage{FG2025}

\FGfinalcopy 

\IEEEoverridecommandlockouts                              
\usepackage{graphicx}
\usepackage{amsmath,amsfonts }
\usepackage{booktabs}
\usepackage{array}
\usepackage{url}
\usepackage{subcaption}
\usepackage{balance} 
\renewcommand{\thefigure}{\arabic{figure}} 

\def\FGPaperID{} 

\title{\LARGE \bf
IlluSign: Illustrating Sign Language Videos by\\ Leveraging the Attention Mechanism 
}

\author{
    Janna Bruner\\
    Reichman University\\
    janna.bruner@post.runi.ac.il \\
    \and
    Amit Moryossef\\
    University of Zurich\\
    amit@sign.mt\\
    \and
    Lior Wolf \\
    Tel Aviv University \\
    wolf@cs.tau.ac.il\\
    }
    
\begin{document}

\ifFGfinal
\thispagestyle{empty}
\pagestyle{empty}
\else
\author{Anonymous FG2025 submission\\ Paper ID \FGPaperID \\}

\pagestyle{plain}
\fi
\maketitle
\pagestyle{plain}

\begin{abstract}

Sign languages are dynamic visual languages that involve hand gestures, in combination with non-manual elements such as facial expressions. While video recordings of sign language are commonly used for education and documentation, the dynamic nature of signs can make it challenging to study them in detail, especially for new learners and educators. This work aims to convert sign language video footage into static illustrations, which serve as an additional educational resource to complement video content. This process is usually done by an artist, and is therefore quite costly.
We propose a method that illustrates sign language videos by leveraging generative models' ability to understand both the semantic and geometric aspects of images. Our approach focuses on transferring a sketch-like illustration style to video footage of sign language, combining the start and end frames of a sign into a single illustration, and using arrows to highlight the hand's direction and motion.
While many style transfer methods address domain adaptation at varying levels of abstraction, applying a sketch-like style to sign language—especially for hand gestures and facial expressions—poses a significant challenge. To tackle this, we intervene in the denoising process of a diffusion model, injecting style as keys and values into high-resolution attention layers, and fusing geometric information from the image and edges as queries.
For the final illustration, we use the attention mechanism to combine the attention weights from both the start and end illustrations, resulting in a soft combination. Our method offers a cost-effective solution for generating sign language illustrations at inference time, addressing the lack of such resources in educational materials. \footnote{Watermarks, such as the SGB-FSS mark in the SignSwiss dataset \protect\cite{signsuisse}, are ignored in this work.
\\[1em]  
Project Page: \\
\url{https://jbruner23.github.io/IlluSign.github.io/} 
}
\end{abstract}


\section{INTRODUCTION}
\input{figures/figure1_overlay_alternative}

\input{figures/first_and_second_method_overview}
Sign languages are rich and complex forms of communication that rely on manual and non-manual articulations to convey meaning, while also requiring the understanding of subtle nuances. Educational resources typically include online video dictionaries, which demonstrate signs in motion.

While videos are used to teach and document sign languages, they can sometimes lack clarity due to the complexity or length of the gesture sequences, rapid movements, or subtle nuances that vary between signs. To overcome these challenges, static illustrations are often used alongside videos, providing a clear and standardized visual representation of individual signs as a complementary resource \cite{signpuddle, signsuisse, british_sign_dictionary}. These are particularly useful for sign language learners, especially those with limited access to in-person instruction. Moreover, illustrations can serve as a bridge between sign language and written or spoken language, making it easier to create educational materials, dictionaries, and accessible content for the deaf community and beyond. Additionally, these illustrations can help bridge communication gaps in crucial public places such as hospitals, banks, and other service areas, quickly enabling workers to understand how to communicate with deaf individuals.

Creating illustrations for sign language videos is often costly, limiting the availability of visual aids. For example, SignSuisse \cite{signsuisse} commissioned only 1,400 illustrations over a five-year span, at an average cost of 33.4 CHF per illustration. With around 18,000 existing entries and plans to record 10,000 more, this highlights the pressing need for more efficient and scalable solutions to generate high-quality illustrations for sign language resources.

In this work, we aim to bridge this gap by developing a method that converts video frames of sign language into detailed and accurate illustrations {as shown in Fig. \ref{fig:final_output}}. These illustrations will serve as a supplementary resource to enhance understanding, facilitate learning, and support non-signers, learners, and educators alike.

In controllable image generation methods such as ControlNet \cite{zhang2023adding}, an additional network is plugged into the diffusion model \cite{rombach2021highresolution} to integrate the condition image. We demonstrate that conditioning on edge maps can be achieved without architectural modifications or additional training on the condition image, as long as the condition input shares some similarity with the reference style image.
Recent studies \cite{hertz2022prompt, alaluf2023crossimage, hertz2023StyleAligned, liu2024understanding} leverage the attention mechanism, manipulating self-attention or cross-attention layers to edit images in a zero-shot manner. Inspired by these approaches, we intervene in the self-attention layers to refine the style of the generated illustration while preserving structural details.

Illustrations depicting hand gestures in sign language typically convey precise instructions for hand positioning. As a result, many of these illustrations consist of a single image containing two pairs of hands—one representing the start position and the other the end position.
Inpainting methods \cite{rombach2021highresolution, 9880056} struggle to generate such images, as they fall outside the distribution of typical training data. Furthermore, generating anatomically accurate hands remains a challenge for most publicly available models \cite{rombach2021highresolution}, even without the additional complexity of combining multiple hand positions in a single image.
To address this, we propose a method that integrates specific features from two separate images using self-attention layers. One image represents the starting hand position, while the other captures the ending position. This approach enables the combination of both hand poses within a single illustration while maintaining structural coherence and stylistic consistency.

\input{figures/final_with_arrow}
\section{RELATED WORK}
Sign language has been a rich area of research, with prior works focusing on recognition, motion analysis, and visualization. Various studies have employed motion capture systems, such as MediaPipe \cite{mediapipe_holistic} and OpenPose \cite{8765346,simon2017hand}, to track body and hand movements, enabling advancements in gesture recognition. However, translating these motions into detailed and stylized illustrations remains an underexplored area. In this section, we review related works to our task.

\textbf{Style Transfer} has been a widely explored area, with significant advancements from GANs \cite{Karras2018ASG}, autoencoders \cite{park2020swapping}, and diffusion models \cite{ye2023ip-adapter}, \cite{alaluf2023crossimage}, \cite{gao2024styleshot}, \cite{wang2024instantstyle}, \cite{hertz2023StyleAligned}, and inversion-based methods \cite{Zhang_2023_inst}. These approaches focus on transforming the style of an input image to reflect the characteristics of a reference image, enabling artistic and domain-specific adaptations.

Our method builds upon these advancements, leveraging diffusion models to address the unique challenges of sign language illustration. Unlike traditional style transfer methods that prioritize artistic elements such as brushstrokes or textures, our approach emphasizes the preservation of structural and semantic details, including precise hand gestures and motion dynamics. By intervening in the denoising process and injecting features into high-resolution attention layers, our method delivers high-quality illustrations tailored to the specific requirements of sign language representation.

\textbf{Sketch-Based Tasks} focus on generating or manipulating sketch-like representations of objects or scenes, balancing abstraction and detail. Methods like CLIPascene \cite{10378256} and CLIPasso \cite{vinker2022clipasso} use CLIP-based supervision to create semantically accurate, abstract sketches. However, their emphasis on high-level shapes makes them unsuitable for capturing the intricate hand gestures required in sign language illustrations. Edge-detection techniques such as Canny \cite{canny1986edge}, xDOG \cite{winnemoller2012xdog}, and HED \cite{xie15hed} extract structural contours but lack semantic awareness, often failing to capture fine details or perform well in low-contrast scenarios. 

These methods are not suitable for sign language illustration because they either prioritize abstraction over detail or fail to integrate semantic understanding with structural fidelity. Sign language illustrations demand precise geometric representation of hand gestures, motion dynamics, and facial expressions, which these methods do not fully address. In contrast, our approach leverages a diffusion-based framework to achieve both semantic alignment and structural precision.

\textbf{Image Editing} and blending have significantly progressed with the advent of generative models, particularly diffusion-based approaches. SDEdit \cite{meng2022sdedit} enables guided image editing by perturbing images with stochastic differential equations, facilitating semantic and structural modifications. Similarly, Blended Diffusion \cite{Avrahami_2022_CVPR} leverages text-driven diffusion models combined with spatial masking to achieve localized edits while preserving coherence between edited and unedited regions. Differential Diffusion \cite{levin2023differential} introduces per-pixel control, allowing for fine-grained and flexible modifications.

While these methods perform exceptionally well in editing natural images, they fall short when applied to the unique challenges of combining overlayed illustrations, such as the precise alignment of geometric and semantic features. To the best of our knowledge, we are the first to address this specific use case. Our approach overcomes these limitations by enabling the accurate blending of illustrations while preserving structural fidelity and intricate details.

\input{figures/qualitive_comp}

\section{Preliminaries}

\noindent{\em Diffusion Models\quad}    During the training process of a diffusion model \cite{pmlr-v37-sohl-dickstein15}, noise is progressively added to an image over a series of time steps until it becomes indistinguishable random noise. The model then learns to reverse this process, step by step denoising the noisy image until it generates a coherent output. Stable Diffusion \cite{rombach2022latent} employs a U-Net architecture \cite{ronneberger2015u}, which consists of an encoder that captures contextual information, a decoder that generates detailed outputs, and skip connections that link corresponding layers between the encoder and decoder.    Each component of the diffusion model—both the encoder and decoder—includes several transformer blocks \cite{NIPS2017_3f5ee243}. Each transformer block is comprised of multi-head self-attention and cross-attention layers. The self-attention layer gathers contextual information from the input image, while the cross-attention layers incorporate information from text prompts as well.  

During training, the attention mechanism computes attention scores for all input tokens to determine the degree of focus to assign to each token. This is defined using the Keys, Queries, and Values vectors, which are learned during training:
    
\begin{equation}
Attention(Q,K,V) = \text{softmax}\left( \frac{Q\cdot K^T}{\sqrt{d_k}} \right) V\,,
\end{equation}
where \( d_k \) represents the dimension of the key vectors.

\noindent{\em B-Splines\quad} Splines are a common representation for curves, allowing smooth and continuous curves while minimizing oscillations between data points \cite{curves}. B spline \cite{joy_bspline} is a piecewise polynomial curve, defined as a linear combination of basis functions: 
    \begin{equation}
    \mathbf{C}(u) = \sum_{i=0}^{n} N_{i,k}(u) \mathbf{P}_i\,,
    \end{equation}
    where  \(N_{i,k}(u)\) are the basis functions of degree \(k\),
         \(\mathbf{P}_i\) are the control points,
        and \(u\) is a parametric coordinate.

\section{METHODOLOGY}

We adopt a modular pipeline to ensure flexibility, user control, and easy component swaps as new technologies emerge. 
We first perform sign segmentation to identify the stroke boundaries: a `start' frame (just after the action preparation phase) and an `end' frame (just before the retraction phase). We then independently generate stylized illustrations of these two frames. Next, we merge them into a single overlay image, preserving both poses while avoiding clutter. Finally, we track the hand trajectory to overlay directional arrows that highlight the motion between frames. By breaking the workflow into modular steps—segmentation, illustration, composition, and motion—our pipeline remains flexible, allowing users to swap or refine any component as improved methods emerge, while also allowing users to intervene at every step. 

\subsection{Segmentation} \label{segmentation}
Movement phases in signs and co-speech gestures \cite{Kita1997MovementPI} typically include: (1) resting; (2) preparation; (3) stroke; (4) hold; (5) retraction; and (6) resting. Ende et al. \cite{6094592} further notes that continuous gestures may replace retraction with repeated stroke–hold cycles.

For our purposes, however, we only require two frames to illustrate a sign: a `start' frame just after preparation (Phase 2) and an `end' frame just before retraction (Phase 5). We obtain these boundaries using the automatic segmentation approach of Moryossef et al. \cite{moryossef-etal-2023-linguistically}, which provides a suitable proxy for our needs. Specifically, the method reliably identifies stroke onset and offset frames, which align closely with the practical `start' and `end' points in a sign.
For very short sign videos, the model may produce inaccurate results, as it was trained on sequences of signs forming sentences rather than isolated signs. To address this, we manually refine the start and end frames when necessary.

\input{figures/different_styles}
\subsection{Style Transfer}

To generate an illustration we use the following inputs: (i) an image \(I_{\textit{}{img}}\) from a video frame, (ii) a lineart \cite{chan2022drawings} annotation as the edges map \(I_{\textit{}{edges}}\), (iii) and a black-and-white sketch-like style image \(I_{\textit{}{style}}\) from SignPuddle \cite{signpuddle}, a publicly available sign language dictionary.

We first invert all images using $T=100$ denoising steps to their latent representations \(z^i_{\textit{T}}\), where $i \in({img, edges, style})$ and $z_T$ is the final noise representation, using the inversion technique proposed by Huberman et al. \cite{huberman2024edit}.
In standard diffusion inference, the latent code typically starts from random noise and evolves to generate the desired image. However, in our approach, we initialize the output image \(I_{\textit{}{out}}\) using the edges' latent code \(z_{\textit{T}}^{edges}\), providing a strong geometric structure. Without further modifications, this process would result in  \(I_{\textit{}{out}}\) = \(I_{edges}\).
To incorporate style and geometry during the diffusion process, we inject the information at each timestep \textit{t}, where  \(t\in[0,70]\) in the last self-attention  \( 64 \times 64 \) resolution layers of the decoder. Specifically, we define:

\begin{equation}
    Q_{geo} = \gamma Q_{img} + \delta Q_{edges}\,,
    \label{new_Q}
\end{equation}
where  \(Q_{\textit{}{img}}\) and  \(Q_{\textit{}{edges}}\) are the Query feature vectors from the driving image and edges, respectively. By setting  $\gamma$=1 and $\delta$=0.5, we balance the trade-off between the information given by the edge map and the original image.
The style transfer is then applied as:

\begin{equation}
    Attention(Q,K,V) = \text{softmax}\left( \frac{Q_{geo}\cdot K_{style}^T}{\sqrt{d_k}} \right) V_{style}
    \label{attention_eq}
\end{equation}

We apply contrast adjustment with contrast factor $\beta=1.67$, as demonstrated in \cite{alaluf2023crossimage}, to enhance the effectiveness of using query, key, and value (Q, K, V) from different images within self-attention. Additionally, we incorporate Adaptive Instance Normalization (AdaIN) \cite{huang2017adain} to better align the generated image with the style reference image, and the classifier-free guidance \cite{ho2021classifierfree} with guidance scale of 3.5 and a simple prompt such as "a woman" or "a man".

\subsection{Illustrations Overlay}

To create the final illustration, we combine two frames: the start and end of the sign. Our hypothesis is that hand positions change significantly, while the face and body posture remain relatively consistent. By modifying the self-attention queries, we achieve a soft overlay of the illustrations, minimizing distortions in the face while integrating changes in hand positions. 
Although the combined illustration lies outside the distribution of the generative diffusion models, this challenge can be addressed by extracting specific features from two separate images and injecting them directly into the final query representations.

We extract hand masks using Grounded-SAM \cite{ren2024grounded}, which combines SAM \cite{kirillov2023segany} with Grounding DINO \cite{liu2023grounding}, using the prompt ``hands''. SAM sometimes struggles with precise body-part segmentation, and we, therefore, incorporate DensePose \cite{Guler2018DensePose} to extract arm segmentations from video frames. 

The overlay process begins by computing the dissimilar features between the two illustrations and applying a mask to visualize important changes. This process often results in issues such as hand occlusion or incomplete visualizations. To address this, we incorporate additional masks for hands and arms to more clearly highlight the critical elements of hand movement.

{Let $m \in\mathbb{N}^{64 \times64}$ represent a binary mask of the features to inject, \(Q_{\textit{}{1}}\), \(Q_{\textit{}{2}}\) the query features from the start and end illustration, respectively, and \(Q_{out}\) query features of the final illustration. All tensors are of size $k\times64\times 64 \times d$, in our implementation, where $k=8$ represents the number of attention heads, 40 channels each. 

We first compute the cosine similarity between the query features along the channel dimension and then take the average across the attention heads:
   \begin{equation}
        \sum_k m_{\text{sim}}[i,j] = \frac{Q_1[i,j] \cdot Q_2[i,j]}{\|Q_1[i,j]\| \|Q_2[i,j]\|}\, / k,
    \end{equation}
where we use $i,j$ as the spatial indices and $Q_1[i,j], Q_2[i,j]\in\mathbb{R}^{k\times d}$. A  dissimilarity mask is then computed by thresholding:
  \[
m_{\mathrm{dis}}[i,j] = \mathbf{1}\{\,m_{\mathrm{sim}}[i,j]<T\}\,,
\]
where in our implementation we use a threshold that is set to the 0.1 quantile. We then combine the hands and arms masks into final masks \( m_1 \), \( m_2 \) for the start and end illustration, respectively.

The final output \(Q_{\textit{}{out}}\) is generated by another deniosing process with $t \in[0,50]$ timesteps, by the following computation steps:

\begin{equation}
    Q_{out} = Q_1 (1-m_{dis}) + Q_2m_{dis}
    \label{q_out1}
\end{equation}
By this step we inject into $Q_{out}$ the unique features from $Q_1$ and $Q_2$. 
We emphasize the hands from the start illustration based on $m_1$, in case of occlusions:
\begin{equation}
    Q_{out} = Q_{out}(1 - m_1) + Q_1 m_1 
    \label{q_out2}
\end{equation}
And for the final step we repeat for end illustration:
\begin{equation}
    Q_{out} = Q_{out}(1 - m_2) + Q_2 m_2 
    \label{q_out3}
\end{equation}
}

In Fig. \ref{first_and_second_stage_method} we highlight the overview of our method.

\input{figures/ablations_queries}

\subsection{Trajectory Arrows}
For hand motion visualization, we track hand keypoints across video frames using MediaPipe Holistic \cite{mediapipe_holistic}.
{We first extract keyframes marking the start and end of the sign using a sign segmentation model \cite{moryossef-etal-2023-linguistically} and then track hand keypoints between these frames using the MediaPipe model.

Out of the 21 landmarks MediaPipe Holistic extracts per hand, we found that the index finger keypoint is the most relevant for most use cases for sign language, and it is the only one used in our current implementation.

After the temporal segmentation step (Sec. \ref{segmentation}}) Most videos contain between 6 to 12 consecutive data points. We then fit a B-spline curve $C$ to the datapoints, producing a clean representation of hand motion, using the MSE loss. 

The trajectory arrow for the new points $u$ is then drawn using VectorGraphics, ensuring smooth and visually intuitive representations of the hand trajectories. Fig. \ref{fig:with_arrows} shows the final illustrations with our trajectory arrows.

\subsection{Implementation Details}
 Our method is built on Stable Diffusion v1.4 \cite{rombach2021highresolution}, taking as input 5 images: 2 images for each illustration (an image and its corresponding edges for both the start and end frames) and a style image. We recommend using a GPU with at least 15GB of RAM for optimal performance. 

\textbf{Illustrations} - before creating the final illustration, we first ensure that the hand masks from the start and end frames do not overlap.  If overlap occurs, we skip the overlaying stage and draw the motion arrows only on the start sign illustration.

\section{Experiments}

\subsection{Qualitative Evaluation}
In Fig. \ref{fig:qualitative}, we compare our results to the state-of-the-art style transfer methods with our approach: StyleShot with Contour and Lineart conditioning \cite{gao2024styleshot}, InST \cite{Zhang_2023_inst} and Cross-Image Attention \cite{alaluf2023crossimage}. As shown, our method achieves strong results in terms of both hand gesture representation and alignment with the target style.
However, challenging cases, such as low-contrast conditions or hands positioned near the face, can reduce the effectiveness of edge detection models in capturing hand edges. Consequently, the hand edges may appear faint or incomplete in the generated illustrations. In these cases, \( StyleShot_{\text{lineart}} \) sometimes provides more interpretable outputs by including hand shadows. Nevertheless, this approach fails to align with the specific style of the target image.
Additionally, our method demonstrates superior performance in generating details like sleeves and shirts, which are often absent in edge maps and, consequently, omitted in \( StyleShot_{\text{lineart}} \). This highlights the advantage of incorporating style and geometric information in our approach to better capture intricate elements of the illustration.

Additionally, in Fig. \ref{fig:with_arrows}, we compare our generated illustrations of ovleray illustrations to manually drawn illustrations from the dataset \cite{signsuisse}. The results demonstrate that our method produces illustrations comparable to the manual ones and, in some cases, even achieves higher accuracy. Additional examples can be found in the supplamentary.

In Fig. \ref{other_styles}, we demonstrate that our method generalizes well across different domains in the style transfer task. Notably, it outperforms on subjects by effectively capturing both structural outlines and fine details. See the supplementary material for additional qualitative examples. The qualitative evaluation on this is outside the scope of this work.

\subsection{Quantitative Evaluation}
We evaluate our model and other style transfer methods across two main categories: hand gesture representation and style alignment.

\textbf{Style Transfer Evaluation:}

For evaluating style transfer, we use standard metrics from prior work \cite{gatys2015neural, gao2024styleshot, alaluf2023crossimage}, Gram Matrix Distance \cite{gatys2015neural}, CLIPScore and LPIPS \cite{zhang2018perceptual} metrics to evaluate the sketch-like style. 
Our ground truth illustrations may include both hands and a directional arrows in orange color, and also the GT illustrations may differ from the person performing the signs in the video frames. To account for this, we first remove the arrows from the illustrations, and then we compare each GT illustration to both the start and end generated illustrations.

\input{tables/evaluation_metrics_style}

\textbf{Hand Gesture Evaluation:}

To assess hand gesture accuracy, we evaluate the mean Intersection over Union (mIOU) of hand masks against our baselines. We extract left and right hand masks separately using Grounded-SAM \cite{ren2024grounded} with prompts such as "left hand" and "right hand".

For each baseline, we compare the predicted masks to the ground truth (GT) masks, which are generated from the original video frames. We exclude the AdaAttN \cite{liu2021adaattn} baseline from our evaluation, as its generated masks were not detected reliably.

\begin{table}[tb]
\caption{\textbf{Mean Intersection Over Union (mIOU).} 
The best and second-best values are marked in \textbf{bold} and \underline{underlined}, respectively.}
\label{miou_metric}  

\centering
\begin{tabular}{lccc}
\toprule
Method & {\small Right Hand $\uparrow$} & {\small Left Hand $\uparrow$} & {\small Overall $\uparrow$} \\
\midrule
Our method & \textbf{0.66} & \textbf{0.68} & \textbf{0.68}  \\
AdaAttn~\cite{liu2021adaattn} & - & - & -  \\
Cross-Image~\cite{alaluf2023crossimage}  &0.48  & 0.51 & 0.50 \\
StyleShot\_Lineart~\cite{wang2024instantstyle} &  \underline{0.64} & \underline{0.68} &  \underline{0.66} \\
StyleShot\_Contour~\cite{gao2024styleshot} & 0.56 & 0.60 & 0.58 \\
InST~\cite{gao2024styleshot} & 0.64 & 0.64 & 0.64  \\
\bottomrule
\end{tabular}
\end{table}

\textbf{Results}:
The evaluation results for style transfer are presented in Table \ref{style_metric}. Our method achieves the best performance, matching Cross-Image \cite{alaluf2023crossimage} on LPIPS and Gram distance metrics while outperforming all baselines on CLIPScore.

For hand gesture accuracy, we summarize the mIOU results in Table~\ref{miou_metric}. Our method achieves the highest mIOU for both hands individually and overall.

Given that Gram Matrix distance is widely used as a loss term in training style transfer models, we primarily rely on this metric for evaluating style alignment.

\input{figures/masks_overlay} 

\textbf{User study}
We conducted a user study to compare our method with two baseline style transfer methods. A total of 20 videos were randomly selected from our dataset, showcasing various hand gestures in different positions, including challenging cases such as motion blur and low contrast. The 58 participants  were asked to evaluate the generated images based on two criteria: style alignment and clarity of hand gesture representation. 

The results, summarized in Table~\ref{tab:user_study}, demonstrate that our method achieved the best overall performance. It received the highest score on the majority of the data, with the exception of one image. 
\begin{table}[t]
\centering 
\caption{\textbf{User Study Results}} 
\label{tab:user_study}  
\begin{tabular}{lc}
\toprule
Method & Overall Quality \\ 
\midrule
Ours & 66.4\% \\
Cross-Image~\cite{alaluf2023crossimage} & 12.7\% \\  
StyleShot\textsubscript{lineart}~\cite{gao2024styleshot} & 20.9\% \\
\bottomrule  
\end{tabular}
\end{table}

\input{figures/ablation_inputs}

\subsection{Ablation Study}
We conducted an ablation study to justify our design choices, particularly our method for combining illustrations and input strategies. 

\textbf{Queries Initialization For Overlay\quad} As detailed in {Eq.~\ref{q_out1}, \ref{q_out2} and \ref{q_out3}}, we compute \(Q_{out}\), representing the combined illustration. Instead of initializing with random noise, we invert the two illustrations and set the latent code {$z^1_T$ of the start sign illustrarion $I_1$ as the initial noise $z^{out}_T$ of the output illustration.} The comparative analysis in Fig. \ref{fig:queries_init}  suggests that, for each timestep \textit{t}, it is beneficial to inject {\(Q_{1}\), \(Q_{2}\) of the original inverted latent codes $z^1_t, z^2_t$ for $t\in[0,T]$}, rather than directly injecting  \(Q_{2}\) into \(Q_{out}\). Omitting the injection of \(Q_{1}\) leads to a degradation in hand accuracy, highlighting the importance of properly initializing and preserving  \(Q_{1}\) features in the combined output. This approach ensures that \(Q_{out}\) preserves the critical features of the illustrations and does not overly rely on the diffusion process. Relying solely on the diffusion could introduce artifacts, as these combined images fall outside the model's training distribution.

\textbf{Overlay Approaches\quad} To create the overlay of the start and end sign illustrations, we explored various approaches. As shown in Fig. \ref{fig:masks}, simple blending in pixel space introduces artifacts such as duplicate features (e.g., double eyebrows or eyes), which can be mitigated by performing the overlay during attention computation. In panel (c), we examined whether inserting hands from the end sign illustration using a mask could suffice. This approach often results in ``floating hands'' due to the dependency on the quality of body-part segmentation and SAM. In panel; (d), we demonstrate how this can be improved by computing a similarity mask between the query features.

We also evaluated our method with and without the application of hand masks. By comparing (f) to (e), we observe that applying the hand masks to the query features significantly enhances the clarity of the final output. Additionally, we improved style alignment by applying a few additional diffusion steps without recomputing the composition.

\textbf{Input Queries\quad} In Equation (3), we show the use of two types of input queries: 
\(Q_{img}\) (from the original frame) and  \(Q_{edges}\) (from the edge map).

In our analysis, we compare the outputs generated using only \(Q_{img}\) or \(Q_{edges}\) as input. As shown in Fig. \ref{fig:ablation_input}, using  \(Q_{img}\) alone as the driving query can result in missing features, as the similarity between the Query and Key (image and style) depends not only on structural alignment but also on color. For example, this can lead to missing fingers that should otherwise be easily distinguishable. Conversely, edge maps are more color-agnostic and align better with our style image in terms of structure, making it easier to locate the correct Key from the style image.

In Fig. \ref{fig:ablation_input}, we demonstrate that while edge maps as input generate more precise structural elements such as hands, they often omit finer details like sleeves and shirts, which are not present in the edge map. By using a linear combination of  \(Q_{img}\) and \(Q_{edges}\) as input, the generated illustrations achieve a better balance, capturing both the overall structure and finer details effectively.

\section{CONCLUSION AND FUTURE WORK}
We present a novel method for generating illustrations from sign language videos without the need for additional training, ensuring both semantic and structural fidelity in the final illustrations.
 Unlike existing methods such as edge detection, sketch generation, or style transfer, our approach emphasizes capturing the fine-grained details of hand gestures, motion dynamics, and facial expressions, which are essential for accurate sign language communication. 
 
Our method successfully generates sketch-like illustrations in a zero-shot inference setting, without requiring additional training. However, further development is needed to enhance its robustness in capturing larger movements of other body parts, such as head and body posture, especially when dealing with non-rigid transformations.

\section*{ACKNOWLEDGMENTS}
We would like to thank Hila Chefer, Rinon Gal, Shimon Malnick, Roni Paiss, Tal Shaharabany, Ben Vardi and Yael Vinker for their insightful feedback.

\clearpage
\addtolength{\textheight}{-3cm}   

\balance
{\small
\bibliographystyle{ieee}
\bibliography{egbib}
}

\end{document}

%% file: figures/figure1_overlay_alternative.tex
\begin{figure}[t] %
    \centering
    \begin{subfigure}{0.24\columnwidth} 
    \includegraphics[width=\linewidth]{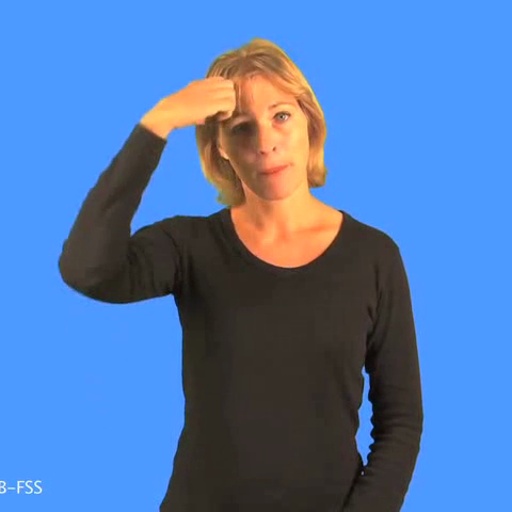}
    \end{subfigure}
    \centering
    \begin{subfigure}{0.24\columnwidth}
        \includegraphics[width=\linewidth]{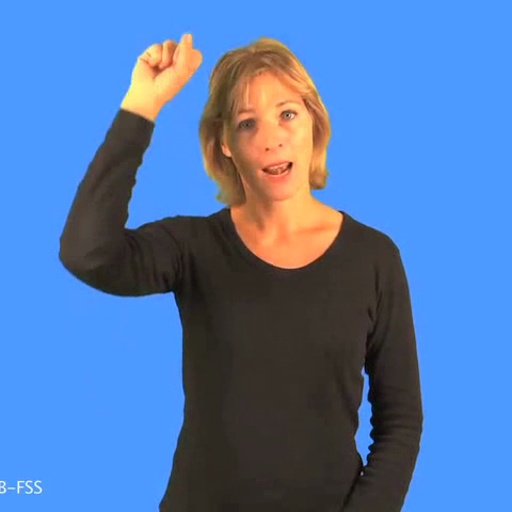}
    \end{subfigure}
    \centering
    \begin{subfigure}{0.24\columnwidth}
        \includegraphics[width=\linewidth]{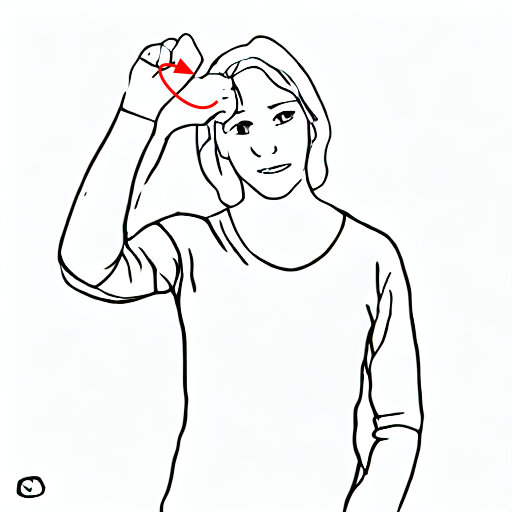}
    \end{subfigure}
    \centering
    \begin{subfigure}{0.24\columnwidth}
        \includegraphics[width=\linewidth]{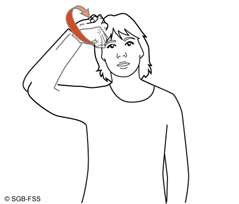}
    \end{subfigure}
    
    \centering
    \begin{subfigure}{0.24\columnwidth} 
    \includegraphics[width=\linewidth]{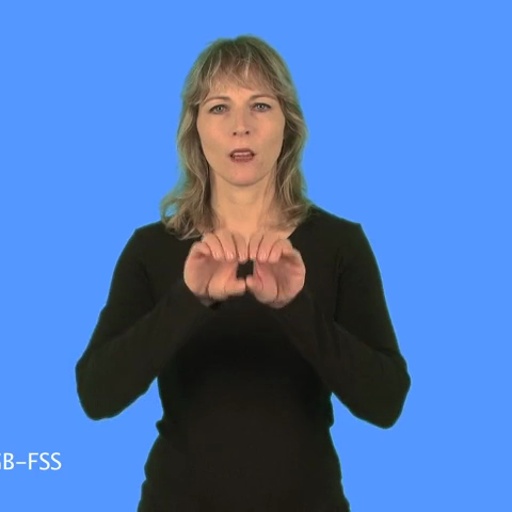}
    \centering
    \end{subfigure}
        \begin{subfigure}{0.24\columnwidth}
        \includegraphics[width=\linewidth]{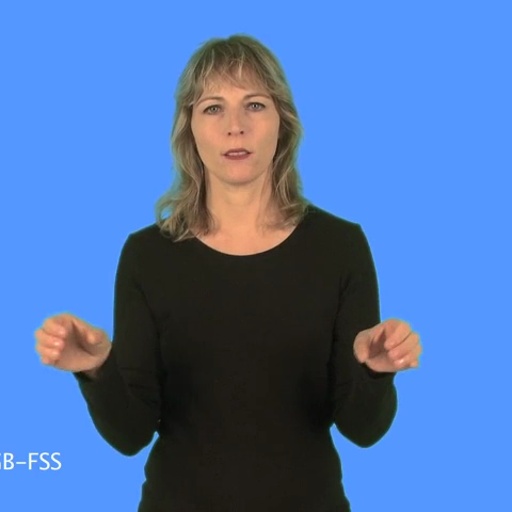}
    \end{subfigure}
    \centering
    \begin{subfigure}{0.24\columnwidth}
        \includegraphics[width=\linewidth]{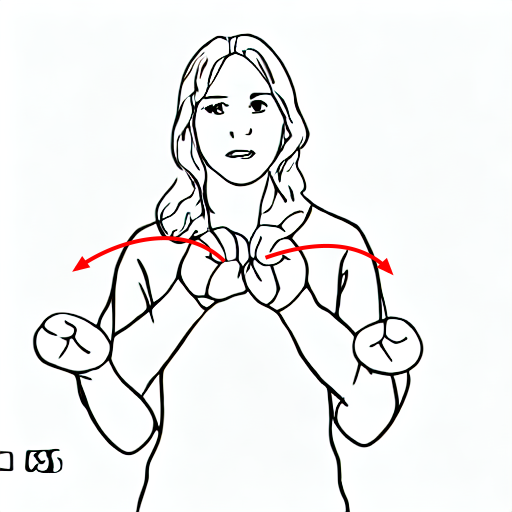}
    \end{subfigure}
    \centering
    \begin{subfigure}{0.24\columnwidth}
        \includegraphics[width=\linewidth]{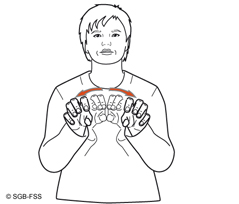}
    \end{subfigure}
    
    \centering
    \begin{subfigure}{0.24\columnwidth} 
    \includegraphics[width=\linewidth]{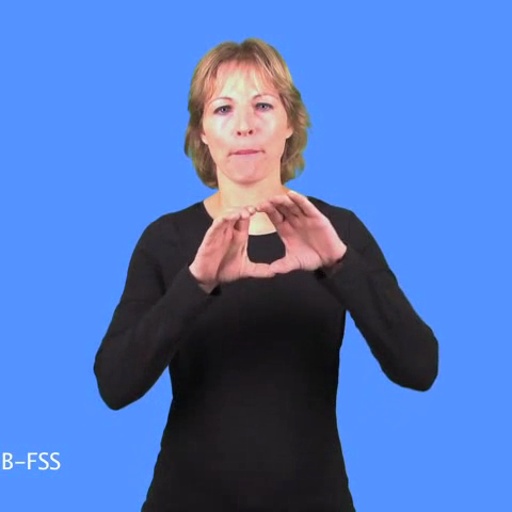}
    \centering
    \par \text{Start Frame}
    \end{subfigure}
    \centering
    \begin{subfigure}{0.24\columnwidth}
    \includegraphics[width=\linewidth]{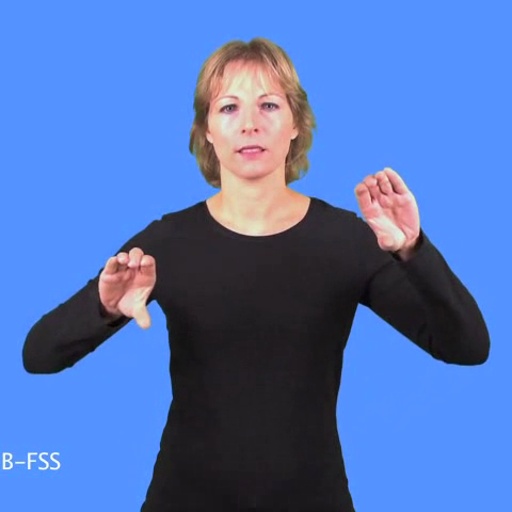}
    \centering
    \par \text{End Frame}
    \end{subfigure}
    \centering
    \begin{subfigure}{0.24\columnwidth}
        \includegraphics[width=\linewidth]{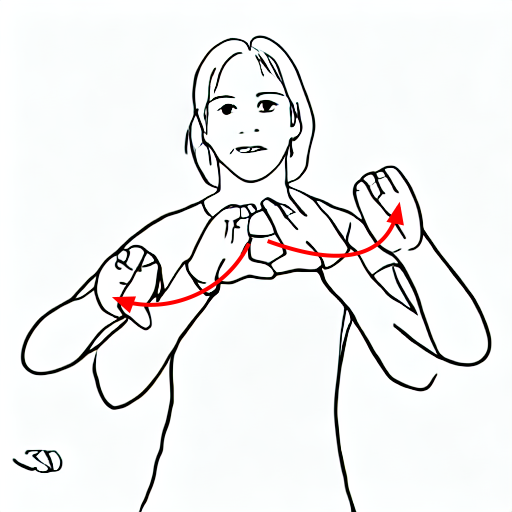}
    \centering
    \par \text{Our Illustration}
    \end{subfigure}
    \centering
    \begin{subfigure}{0.24\columnwidth}
        \includegraphics[width=\linewidth]{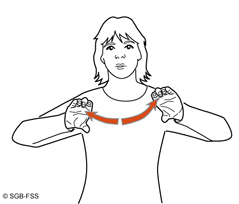}
    \centering
    \par \text{GT}
    \end{subfigure}
    
    \caption{In this work, we present a method for transforming sign language video frames into illustrations that capture the geometric details of the images, emphasizing hand gestures, direction and motion.}
    \label{fig:final_output} 
\end{figure}

%% file: figures/first_and_second_method_overview.tex
\begin{figure*}[tb] %
    \centering
    \includegraphics[width=\textwidth]{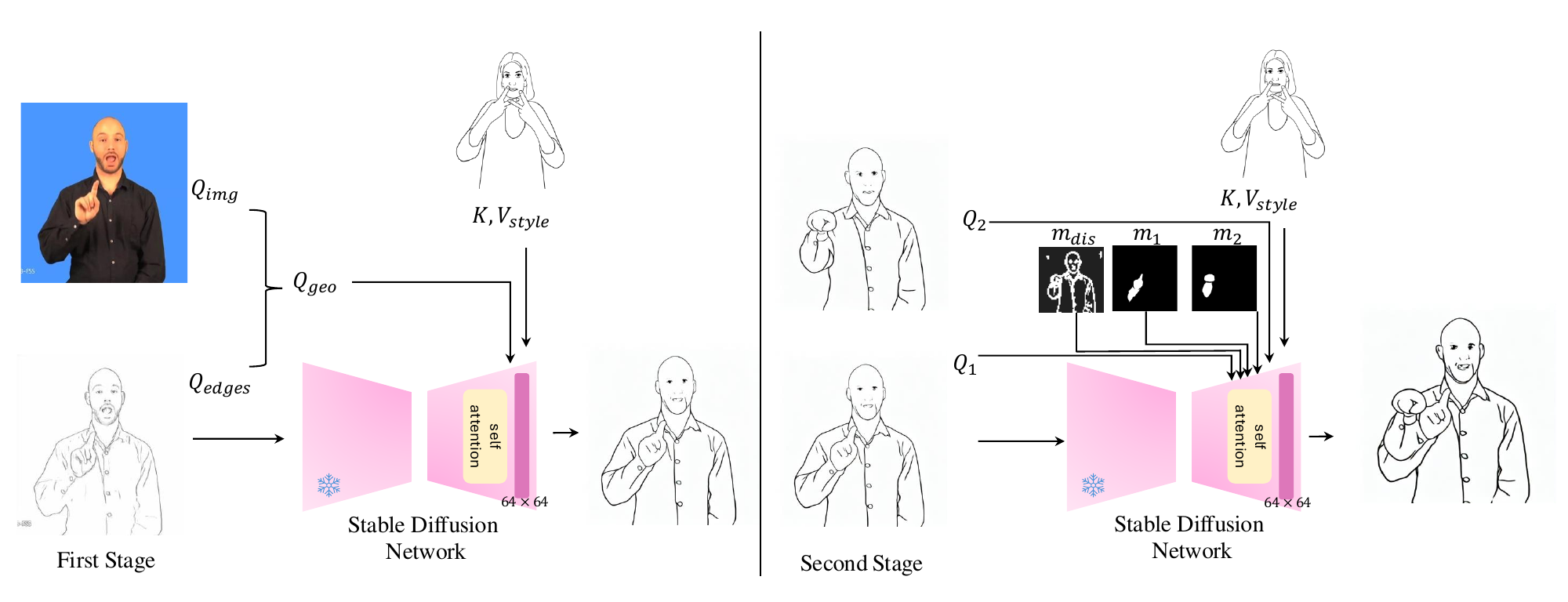}
    \caption{\textbf{Method Overview}: In the first stage we start by inverting images to their latent noise representation and initiate the diffusion process from the edges image noise. In the final resolution attention layers of the decoder, we inject the Keys and Values from the style image, and the Queries as a linear combination of the queries derived from the image and the queries from the edge map.
    In the second stage we apply another diffusion process to fuse query features from the start image and end image, initializing with the latent noise of the start sign image. In the last resolution attention layers, we inject into the queries a combination of unsimilar features and hand masks. The unsimilar features between the queries contribute to the soft overlay of the images, while the hand masks enhance the appearance of the hands.}
    \label{first_and_second_stage_method}
\end{figure*}

%% file: figures/final_with_arrow.tex
\begin{figure*}[t] %
    \centering
    \begin{subfigure}{0.28\columnwidth} %
        \includegraphics[width=\linewidth]{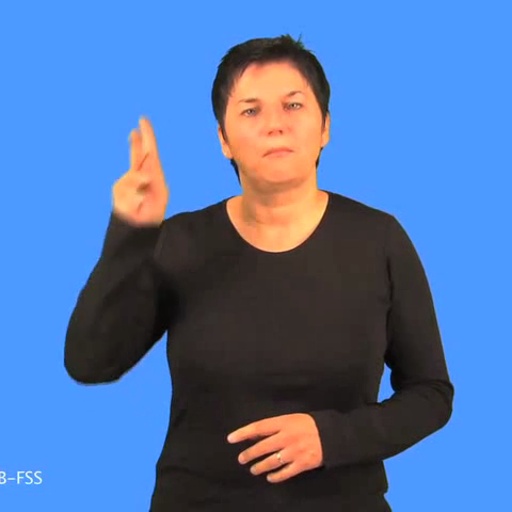}
    \end{subfigure}
        \centering
    \begin{subfigure}{0.28\columnwidth} %
        \includegraphics[width=\linewidth]{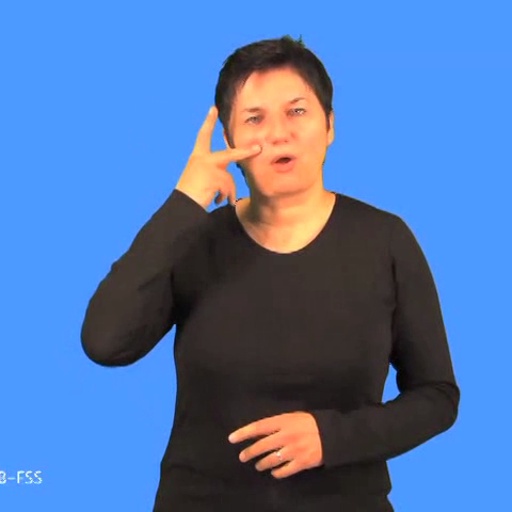}
    \end{subfigure}
    \centering
    \begin{subfigure}{0.28\columnwidth} %
        \includegraphics[width=\linewidth]{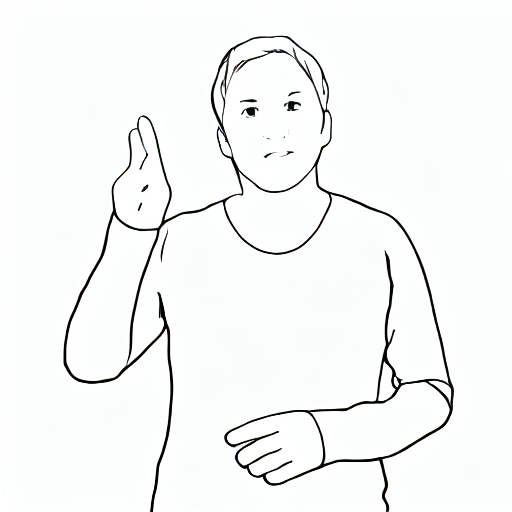}
    \end{subfigure}
    \begin{subfigure}{0.28\columnwidth}
        \includegraphics[width=\linewidth]{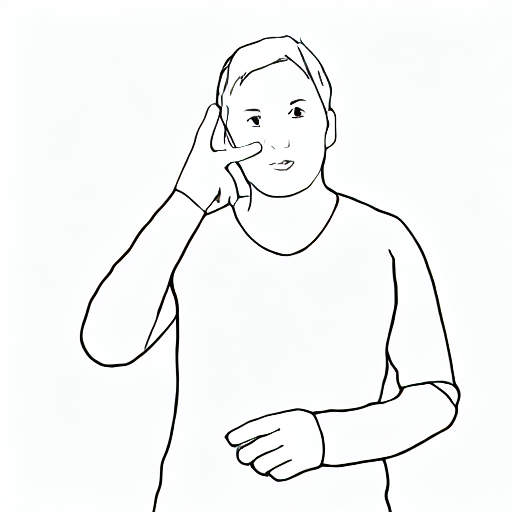}
    \end{subfigure}
        \begin{subfigure}{0.28\columnwidth}
        \includegraphics[width=\linewidth]{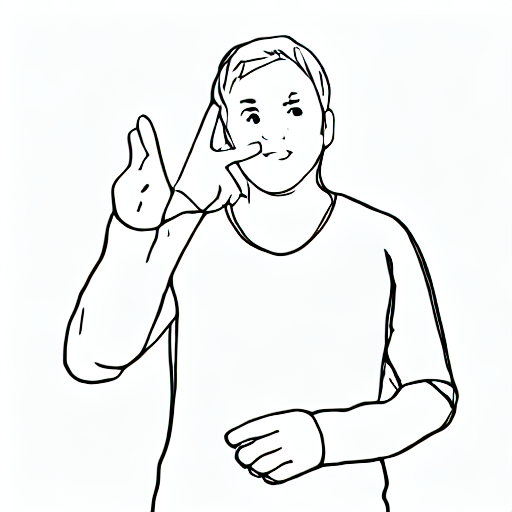}
    \end{subfigure}
    \begin{subfigure}{0.28\columnwidth}
        \includegraphics[width=\linewidth]{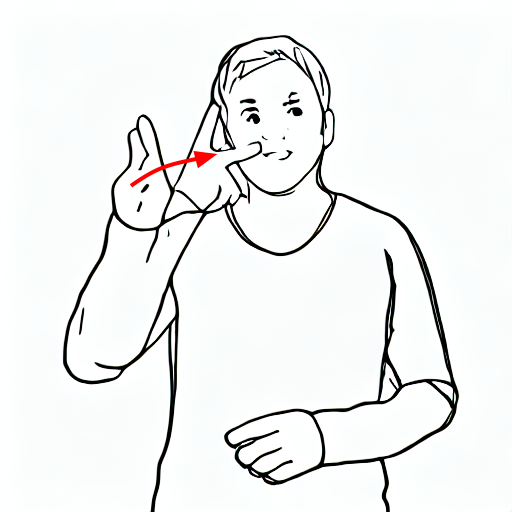}
    \end{subfigure}
    \begin{subfigure}{0.28\columnwidth}
        \includegraphics[width=\linewidth]{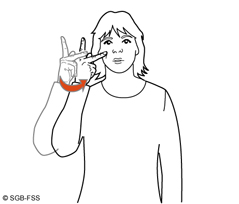}
    \end{subfigure}

    \begin{subfigure}{0.28\columnwidth}
        \includegraphics[width=\linewidth]{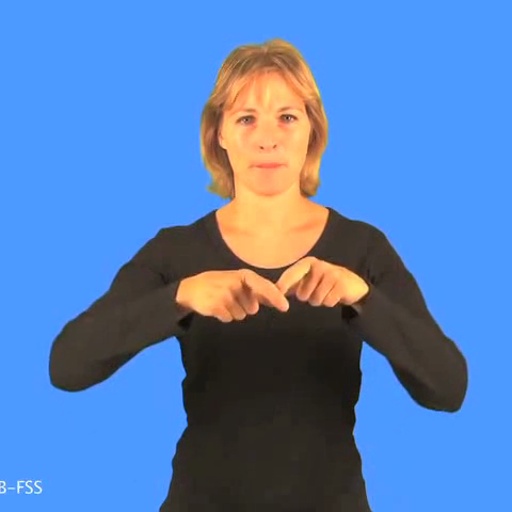}
    \end{subfigure}
        \begin{subfigure}{0.28\columnwidth}
        \includegraphics[width=\linewidth]{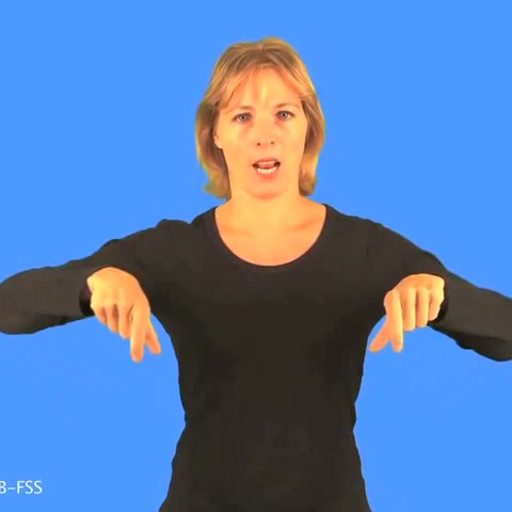}
    \end{subfigure}
    \begin{subfigure}{0.28\columnwidth}
        \includegraphics[width=\linewidth]{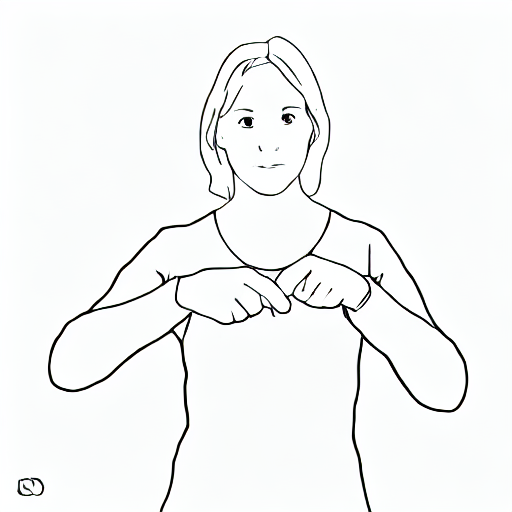}
    \end{subfigure}
    \begin{subfigure}{0.28\columnwidth}
        \includegraphics[width=\linewidth]{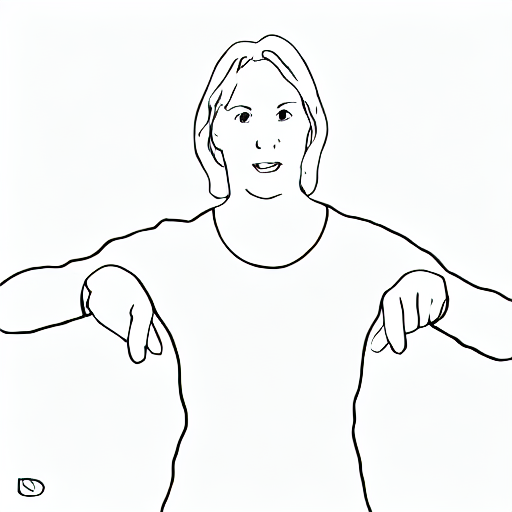}
    \end{subfigure}
    \begin{subfigure}{0.28\columnwidth}
        \includegraphics[width=\linewidth]{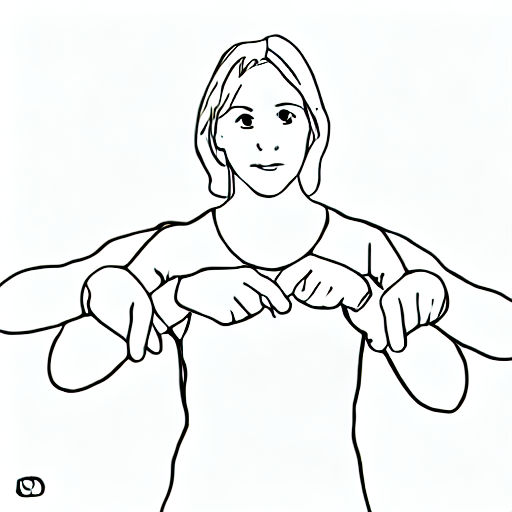}
    \end{subfigure}
    \begin{subfigure}{0.28\columnwidth}
        \includegraphics[width=\linewidth]{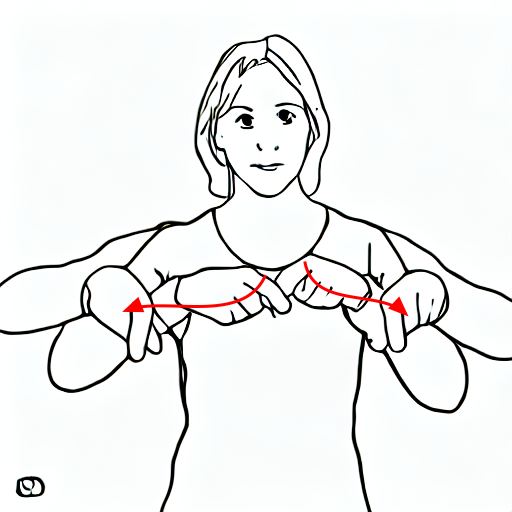}
    \end{subfigure}
    \begin{subfigure}{0.28\columnwidth}
        \includegraphics[width=\linewidth]{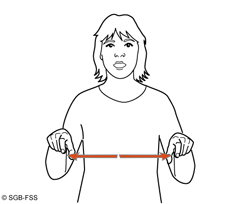}
    \end{subfigure}
    \centering
    \begin{subfigure}{0.28\columnwidth}
        \includegraphics[width=\linewidth]{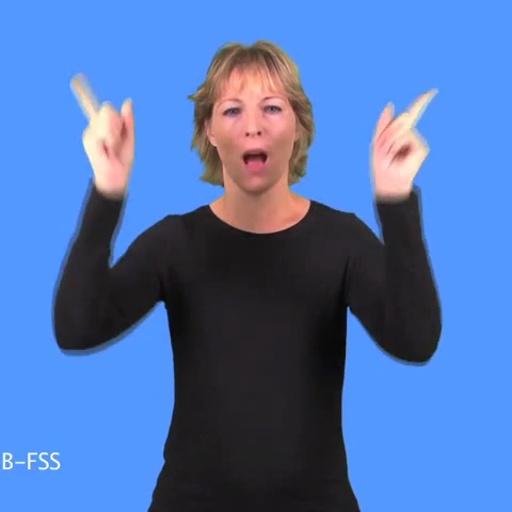}
    \end{subfigure}
    \begin{subfigure}{0.28\columnwidth}
        \includegraphics[width=\linewidth]{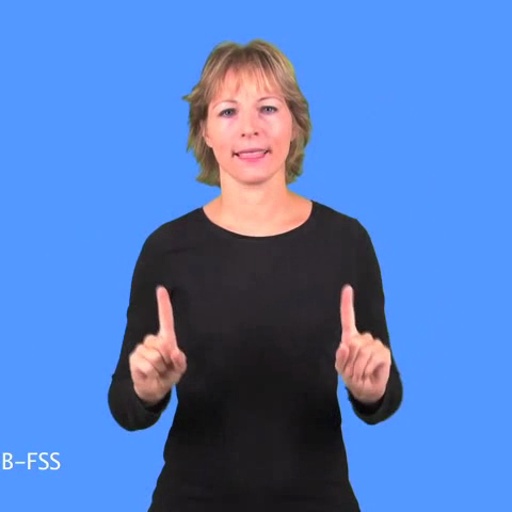}
        \end{subfigure}
    \begin{subfigure}{0.28\columnwidth}
        \includegraphics[width=\linewidth]{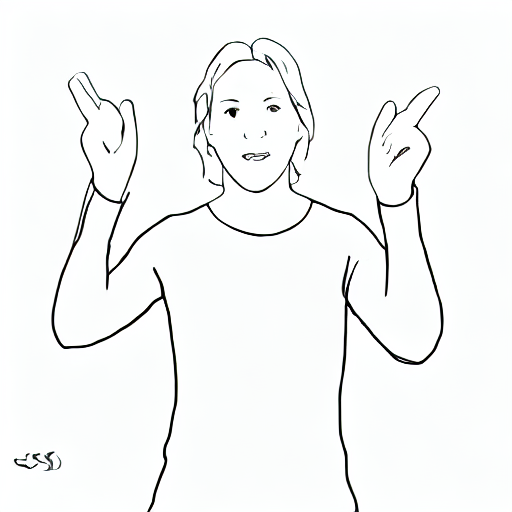}
    \end{subfigure}
    \begin{subfigure}{0.28\columnwidth}
        \includegraphics[width=\linewidth]{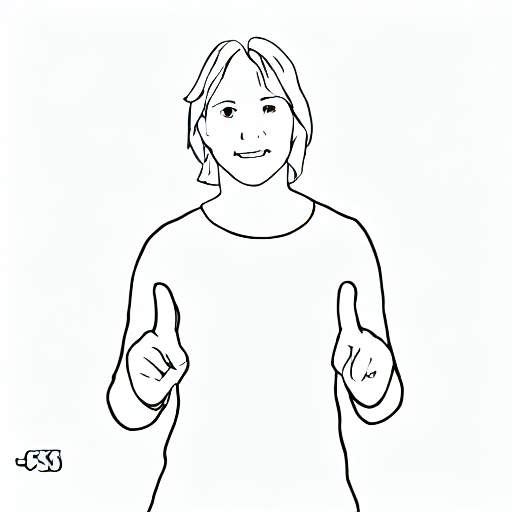}
    \end{subfigure}
        \begin{subfigure}{0.28\columnwidth}
        \includegraphics[width=\linewidth]{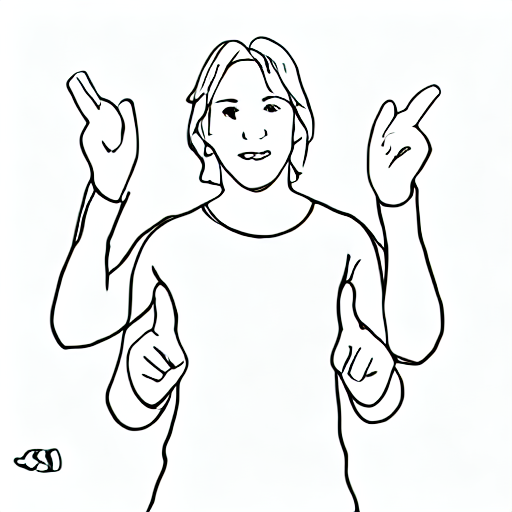}
    \end{subfigure}
    \begin{subfigure}{0.28\columnwidth}
        \includegraphics[width=\linewidth]{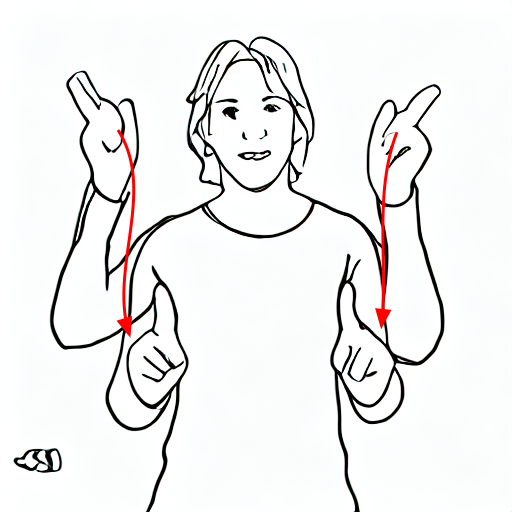}
    \end{subfigure}
        \begin{subfigure}{0.28\columnwidth}
        \includegraphics[width=\linewidth]{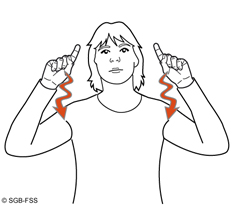}
    \end{subfigure}

    \centering
        \begin{subfigure}{0.28\columnwidth} %
        \includegraphics[width=\linewidth]{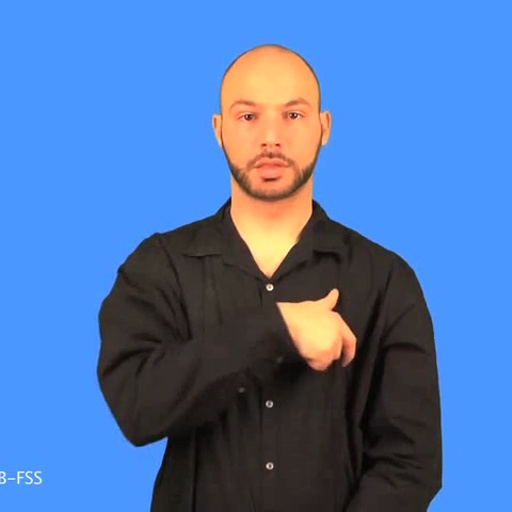}
    \end{subfigure}
    \centering
    \begin{subfigure}{0.28\columnwidth} %
        \includegraphics[width=\linewidth]{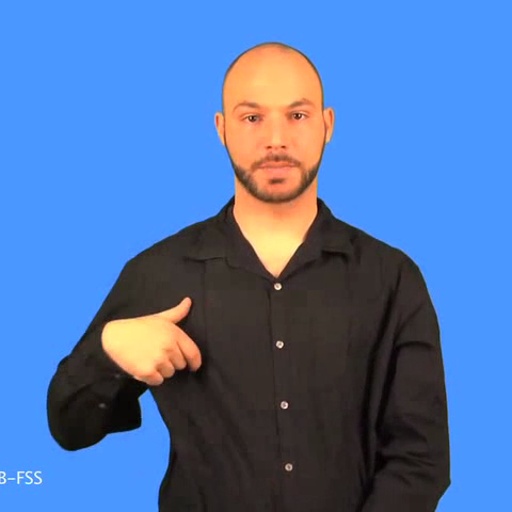}
    \end{subfigure}
    \begin{subfigure}{0.28\columnwidth} %
        \includegraphics[width=\linewidth]{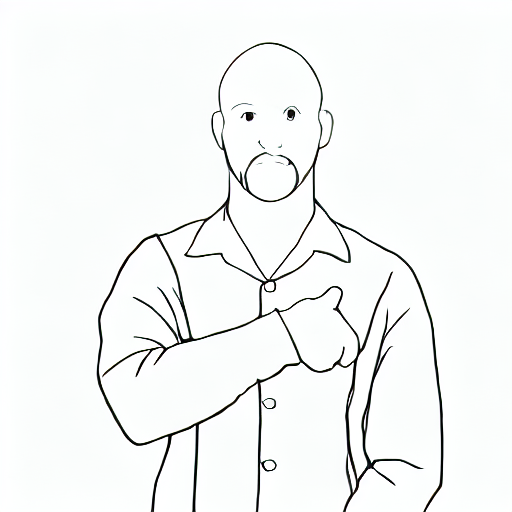}
    \end{subfigure}
        \begin{subfigure}{0.28\columnwidth}
        \includegraphics[width=\linewidth]{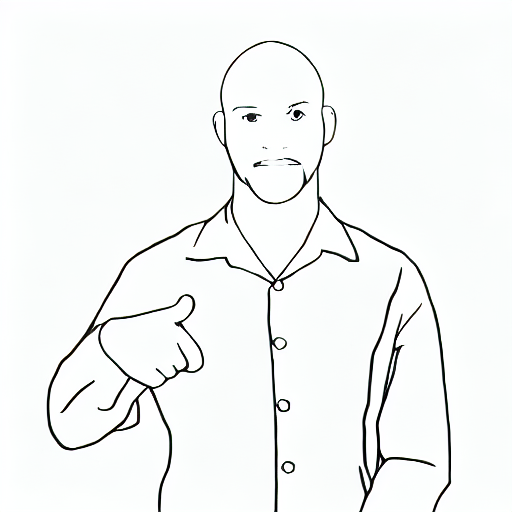}
    \end{subfigure}
    \begin{subfigure}{0.28\columnwidth}
        \includegraphics[width=\linewidth]{figures/method_overlay/name_out_q2---seed_42.png}
    \end{subfigure}
    \begin{subfigure}{0.28\columnwidth}
        \includegraphics[width=\linewidth]{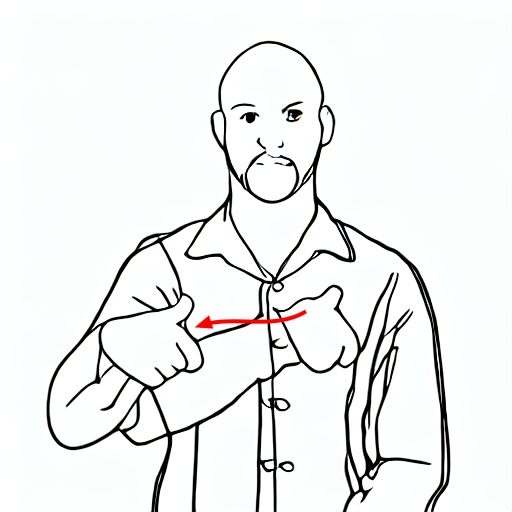}
    \end{subfigure}
        \begin{subfigure}{0.28\columnwidth}
        \includegraphics[width=\linewidth]{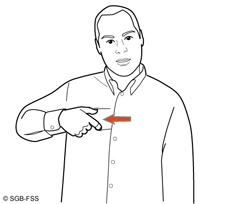}
    \end{subfigure}

    \begin{subfigure}{0.28\columnwidth}
        \includegraphics[width=\linewidth]{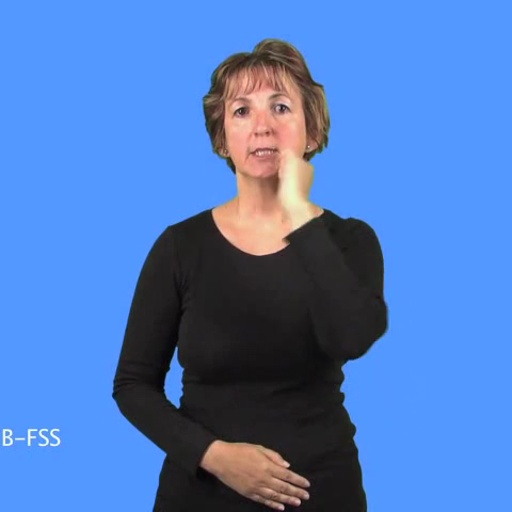}
    \end{subfigure}
    \begin{subfigure}{0.28\columnwidth}
        \includegraphics[width=\linewidth]{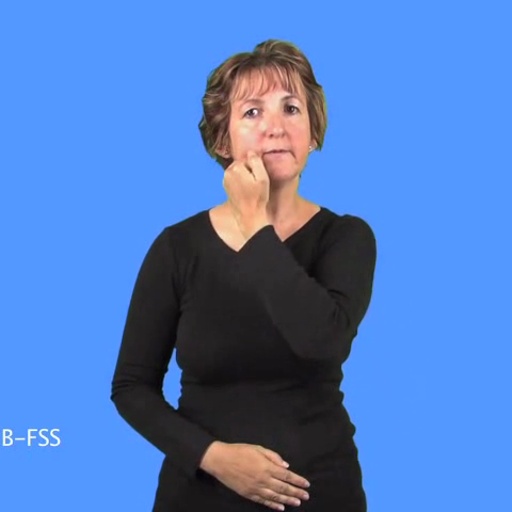}
    \end{subfigure}
    \begin{subfigure}{0.28\columnwidth}
        \includegraphics[width=\linewidth]{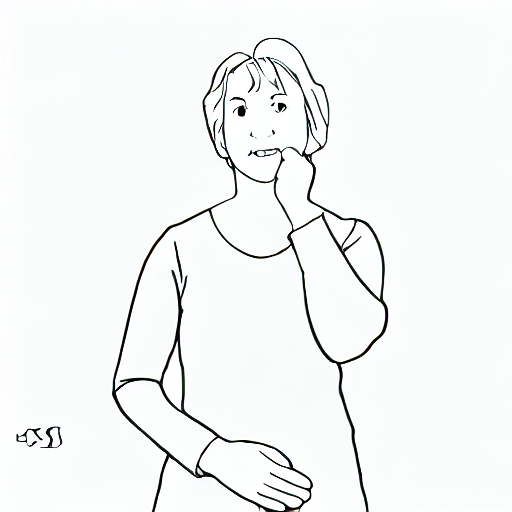}
    \end{subfigure}
    \begin{subfigure}{0.28\columnwidth}
        \includegraphics[width=\linewidth]{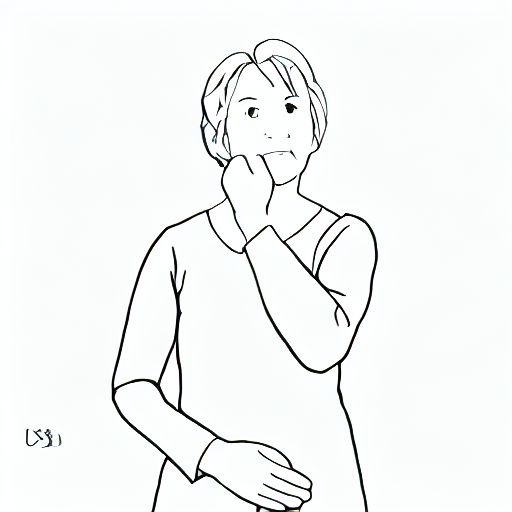}
    \end{subfigure}
    \begin{subfigure}{0.28\columnwidth}
        \includegraphics[width=\linewidth]{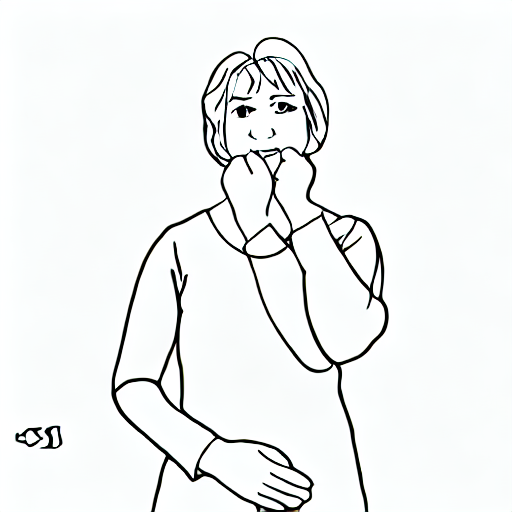}
    \end{subfigure}
    \begin{subfigure}{0.28\columnwidth}
        \includegraphics[width=\linewidth]{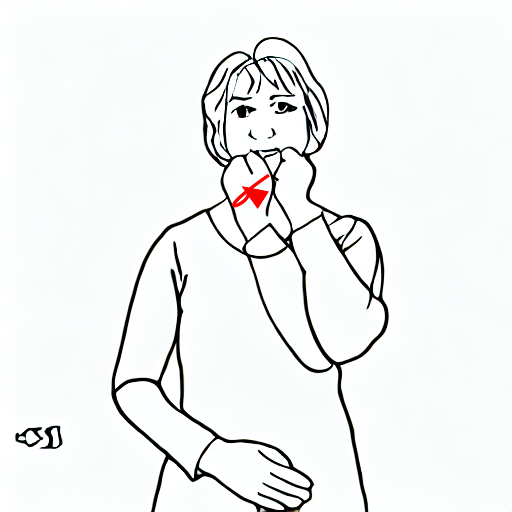}
    \end{subfigure}
    \begin{subfigure}{0.28\columnwidth}
        \includegraphics[width=\linewidth]{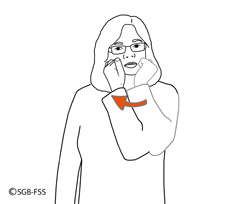}
    \end{subfigure}

    \begin{subfigure}{0.28\columnwidth}
        \includegraphics[width=\linewidth]{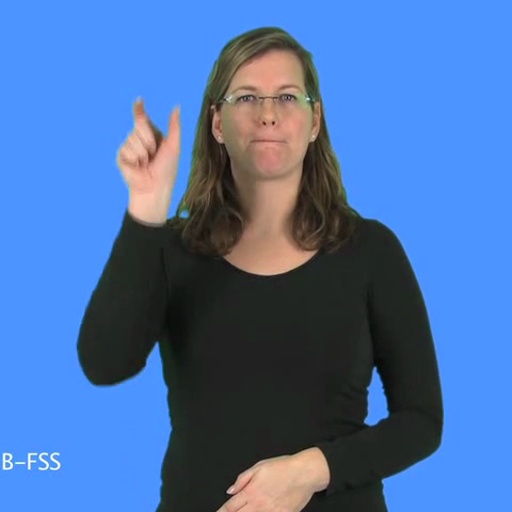}
        \centering
        \text{Start Sign}
    \end{subfigure}
    \begin{subfigure}{0.28\columnwidth}
        \includegraphics[width=\linewidth]{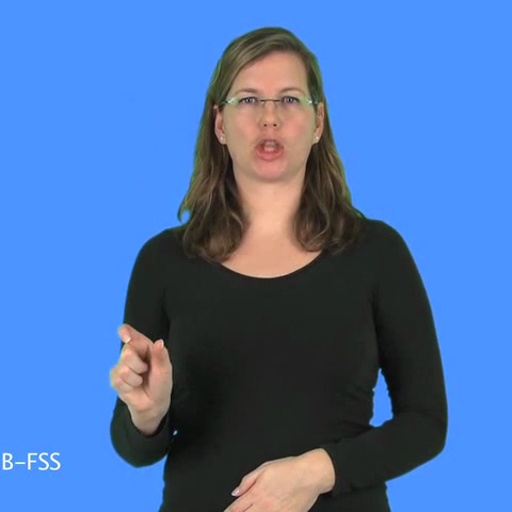}
        \centering
        \text{End Sign}
    \end{subfigure}
    \begin{subfigure}{0.28\columnwidth}
        \includegraphics[width=\linewidth]{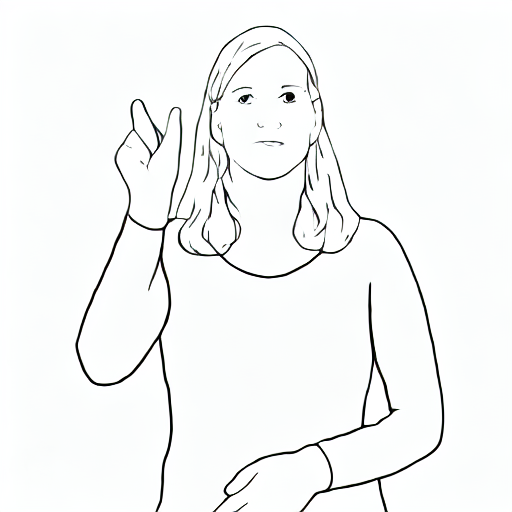}
        \centering
        \text{Start Illustration}
    \end{subfigure}
    \begin{subfigure}{0.28\columnwidth}
        \includegraphics[width=\linewidth]{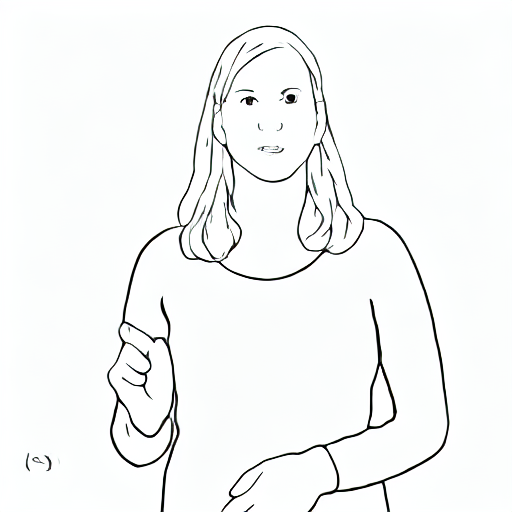}
        \centering
        \text{End Illustration}
    \end{subfigure}
    \begin{subfigure}{0.28\columnwidth}
        \includegraphics[width=\linewidth]{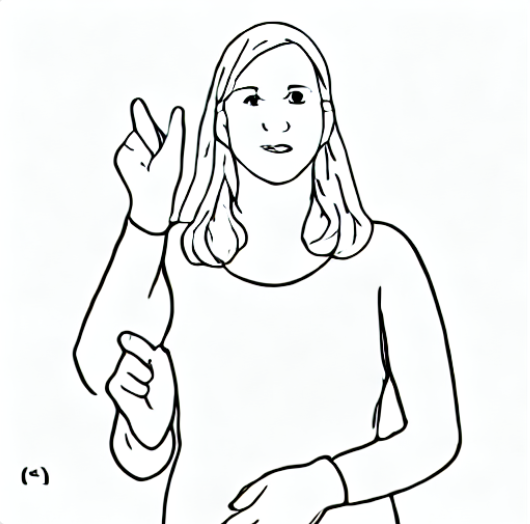}
        \text{Illustration Overlay}
        \end{subfigure}
    \begin{subfigure}{0.28\columnwidth}
    \includegraphics[width=\linewidth]{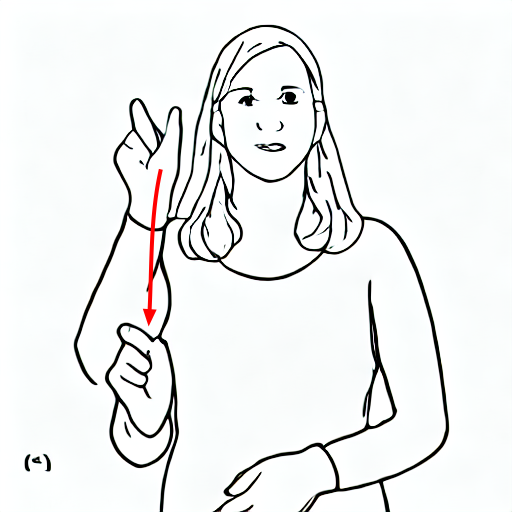}
    \centering
    \text{With Arrow}
    \end{subfigure}
    \begin{subfigure}{0.28\columnwidth}
    \includegraphics[width=\linewidth]{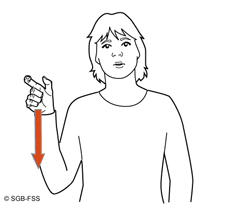}
    \centering
    \text{GT Illustration}
    \end{subfigure}
    
    \caption{Final Illustrations, including the intermediate steps of our method. The process begins with two input frames, which are first transformed into the desired illustration style. Next, an overlay step is applied, followed by the addition of directional arrows. The rightmost column displays the ground truth (GT) illustration from the online dictionary \cite{signsuisse}.}
    \label{fig:with_arrows}
\end{figure*}

%% file: figures/qualitive_comp.tex

\begin{figure*}[t]
    
    \centering
    \begin{minipage}{0.13\textwidth}
        \centering
        \includegraphics[width=\linewidth]{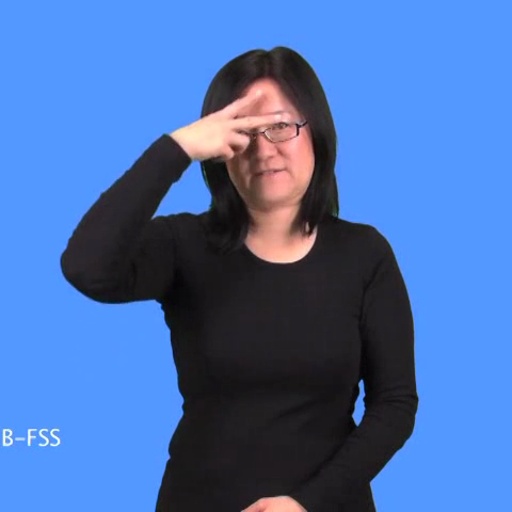}
    \end{minipage}
        \begin{minipage}{0.15\textwidth}
        \centering
        \includegraphics[width=\linewidth]{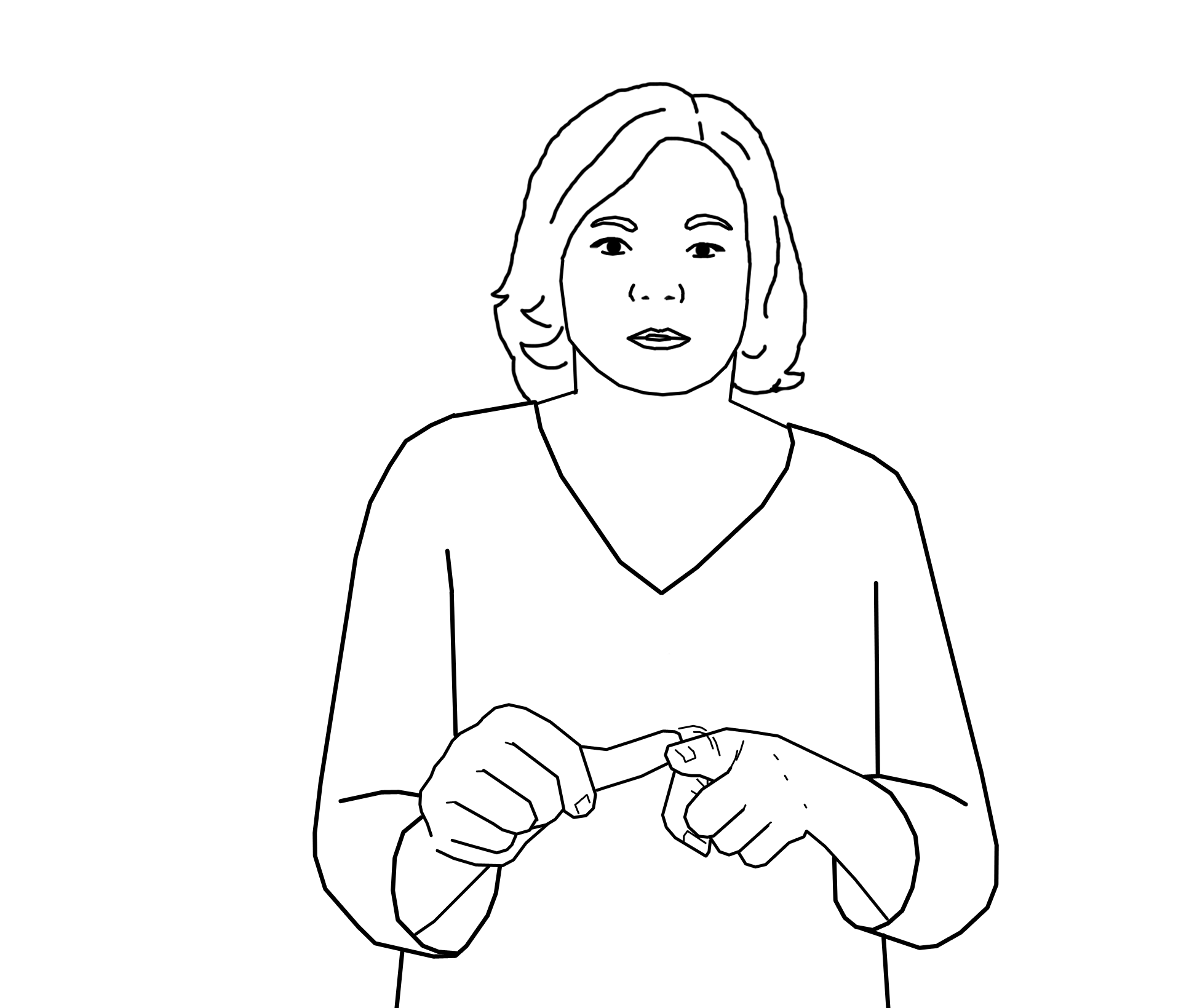}
    \end{minipage}
    \begin{minipage}{0.13\textwidth}
        \centering
        \includegraphics[width=\linewidth]{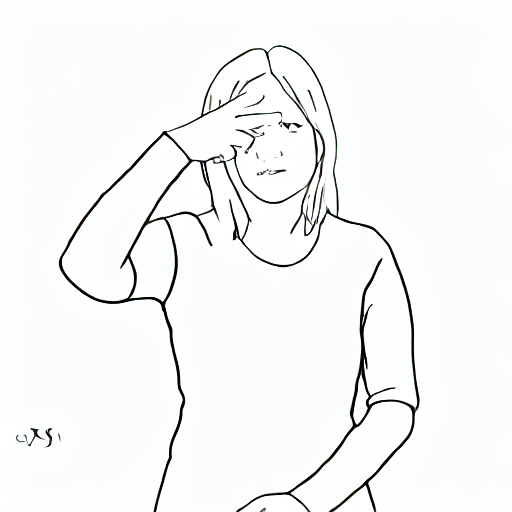}
    \end{minipage}
    \begin{minipage}{0.13\textwidth}
        \centering
        \includegraphics[width=\linewidth]{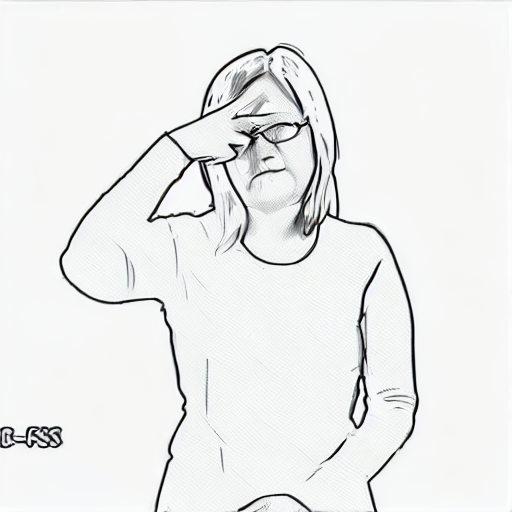}
    \end{minipage}
    \begin{minipage}{0.13\textwidth}
        \centering
        \includegraphics[width=\linewidth]{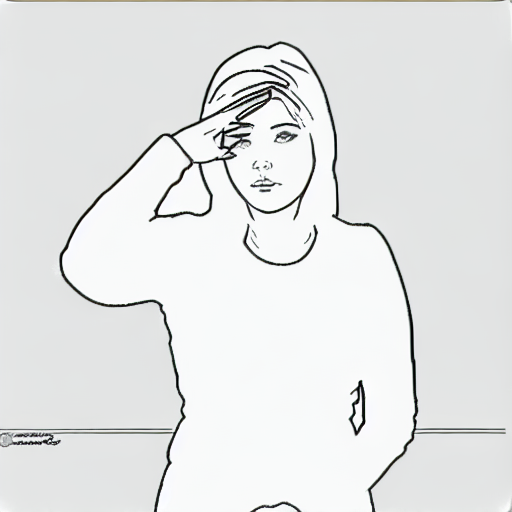}
    \end{minipage}
    \begin{minipage}{0.13\textwidth}
        \centering
        \includegraphics[width=\linewidth]{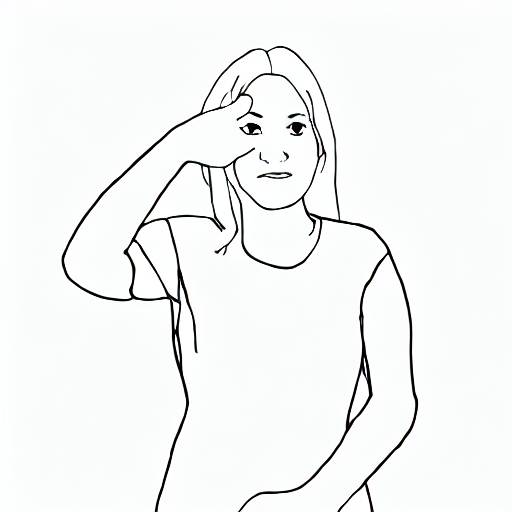}
    \end{minipage}
    \begin{minipage}{0.13\textwidth}
        \centering
        \includegraphics[width=\linewidth]{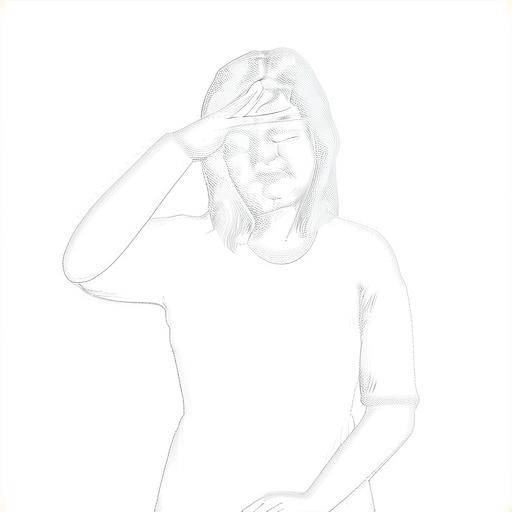}
    \end{minipage}

    \centering
    \begin{minipage}{0.13\textwidth}
        \centering
        \includegraphics[width=\linewidth]{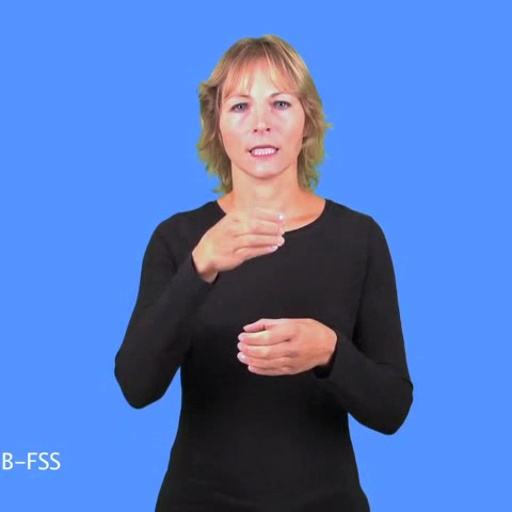}
    \end{minipage}
        \begin{minipage}{0.15\textwidth}
        \centering
        \includegraphics[width=\linewidth]{figures/style_images/10.png}
    \end{minipage}
    \begin{minipage}{0.13\textwidth}
        \centering
        \includegraphics[width=\linewidth]{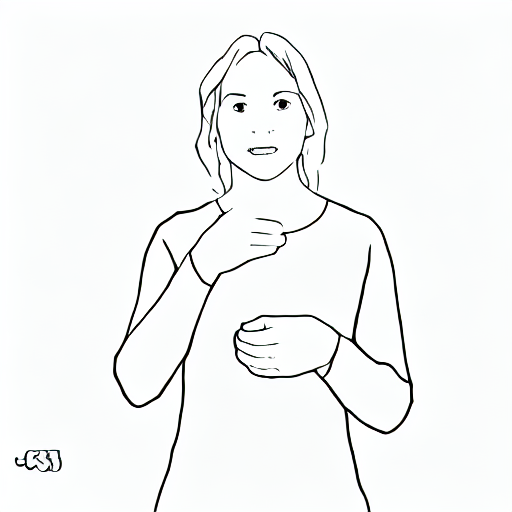}
    \end{minipage}
    \begin{minipage}{0.13\textwidth}
        \centering
        \includegraphics[width=\linewidth]{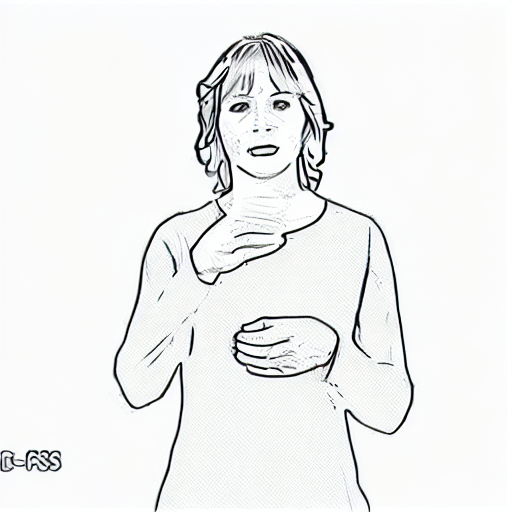}
    \end{minipage}
    \begin{minipage}{0.13\textwidth}
        \centering
        \includegraphics[width=\linewidth]{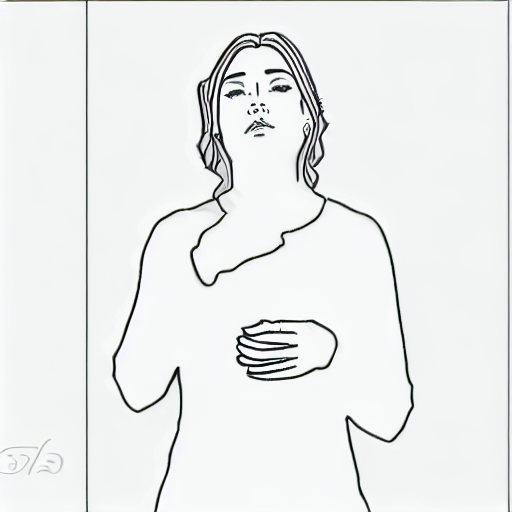}
    \end{minipage}
    \begin{minipage}{0.13\textwidth}
        \centering
        \includegraphics[width=\linewidth]{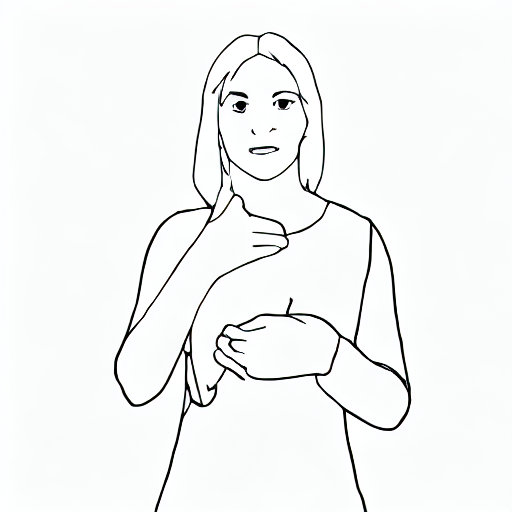}
    \end{minipage}
    \begin{minipage}{0.13\textwidth}
        \centering
        \includegraphics[width=\linewidth]{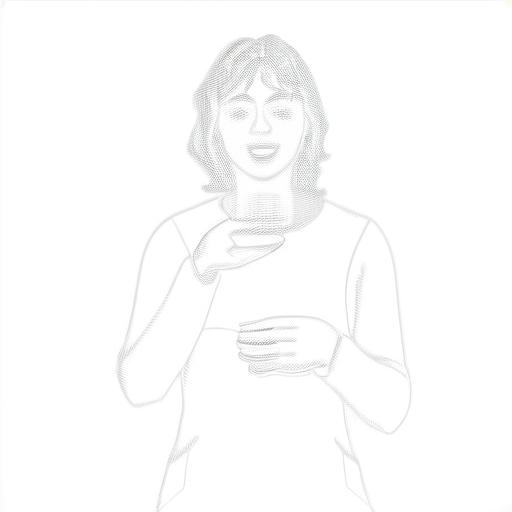}
    \end{minipage}

    \centering
    \begin{minipage}{0.13\textwidth}
        \centering
        \includegraphics[width=\linewidth]{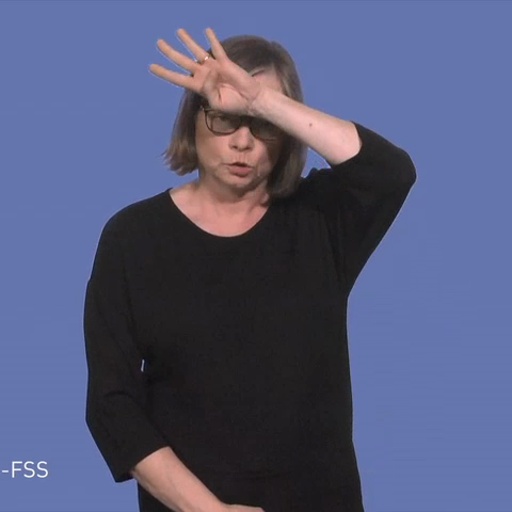}
    \end{minipage}
        \begin{minipage}{0.15\textwidth}
        \centering
        \includegraphics[width=\linewidth]{figures/style_images/10.png}
    \end{minipage}
    \begin{minipage}{0.13\textwidth}
        \centering
        \includegraphics[width=\linewidth]{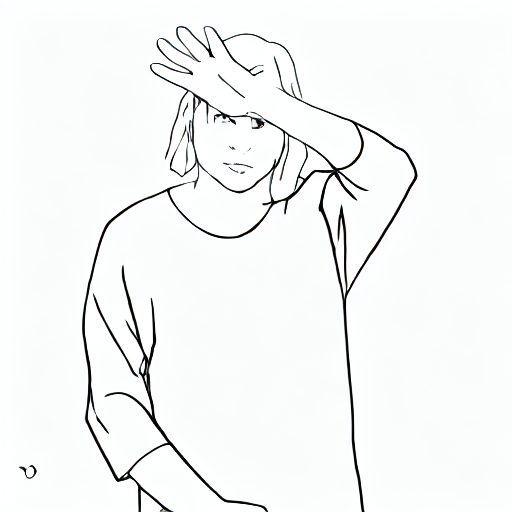}

    \end{minipage}
    \begin{minipage}{0.13\textwidth}
        \centering
        \includegraphics[width=\linewidth]{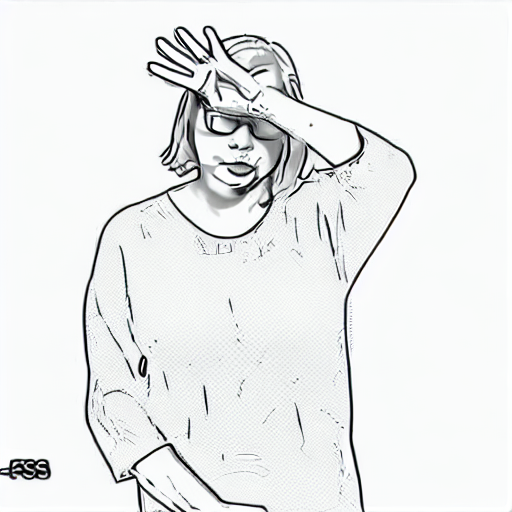}
    \end{minipage}
    \begin{minipage}{0.13\textwidth}
        \centering
        \includegraphics[width=\linewidth]{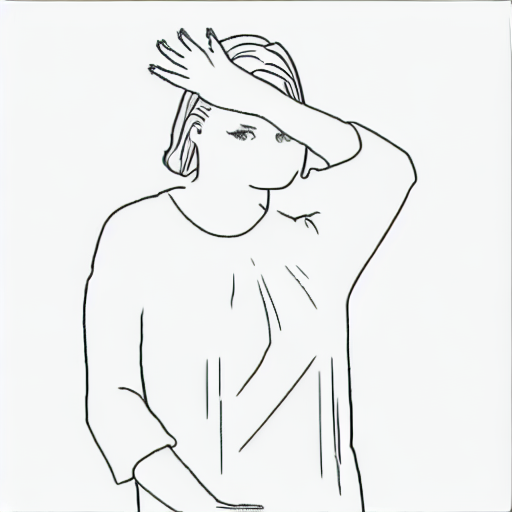}
    \end{minipage}
    \begin{minipage}{0.13\textwidth}
        \centering
        \includegraphics[width=\linewidth]{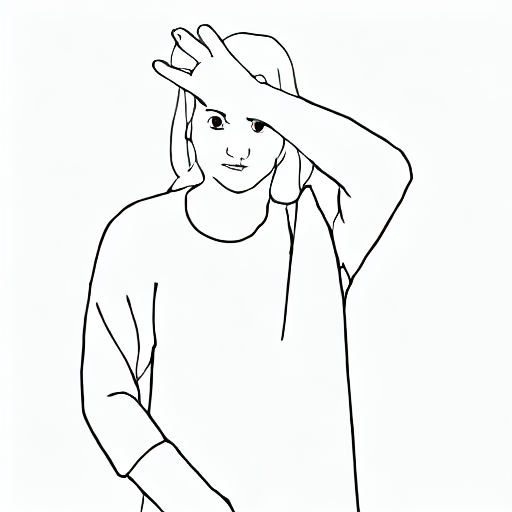}
    \end{minipage}
    \begin{minipage}{0.13\textwidth}
        \centering
        \includegraphics[width=\linewidth]
        {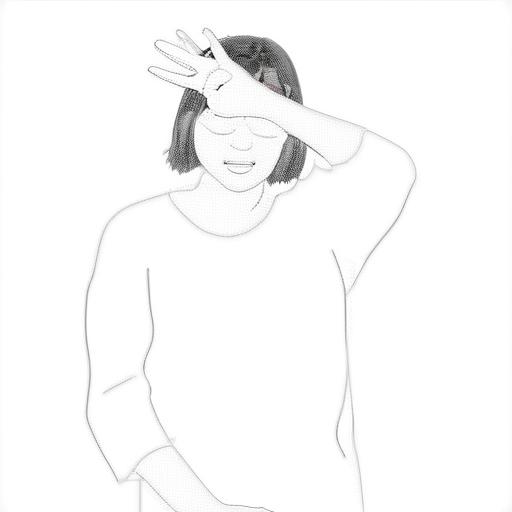}
    \end{minipage}

    \centering
    \begin{minipage}{0.13\textwidth}
        \centering
        \includegraphics[width=\linewidth]{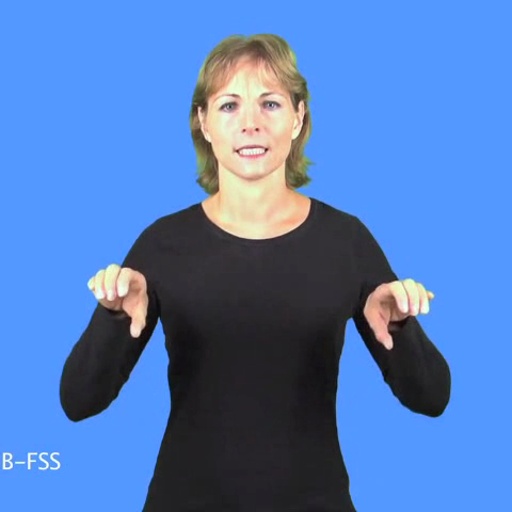}
    \end{minipage}
        \begin{minipage}{0.15\textwidth}
        \centering
        \includegraphics[width=\linewidth]{figures/style_images/10.png}
    \end{minipage}
    \begin{minipage}{0.13\textwidth}
        \centering
        \includegraphics[width=\linewidth]{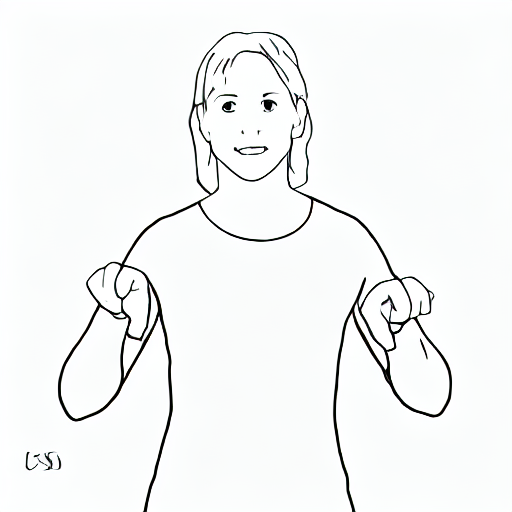}
    \end{minipage}
    \begin{minipage}{0.13\textwidth}
        \centering
        \includegraphics[width=\linewidth]{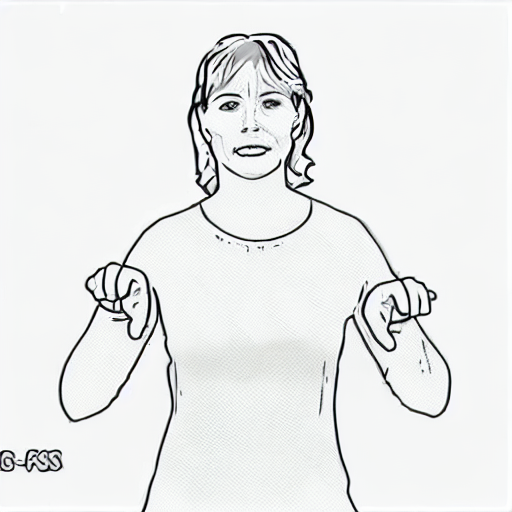}
    \end{minipage}
    \begin{minipage}{0.13\textwidth}
        \centering
        \includegraphics[width=\linewidth]{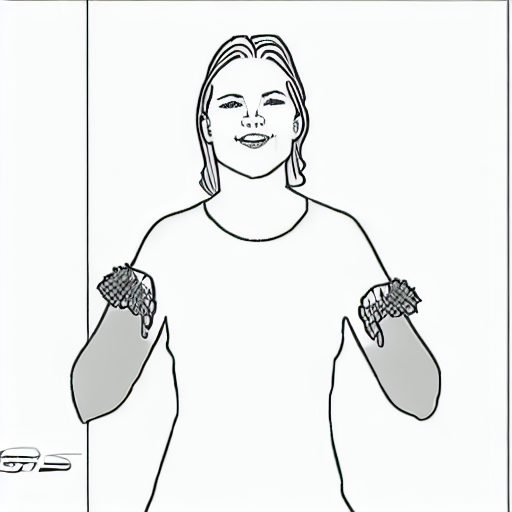}
    \end{minipage}
    \begin{minipage}{0.13\textwidth}
        \centering
        \includegraphics[width=\linewidth]{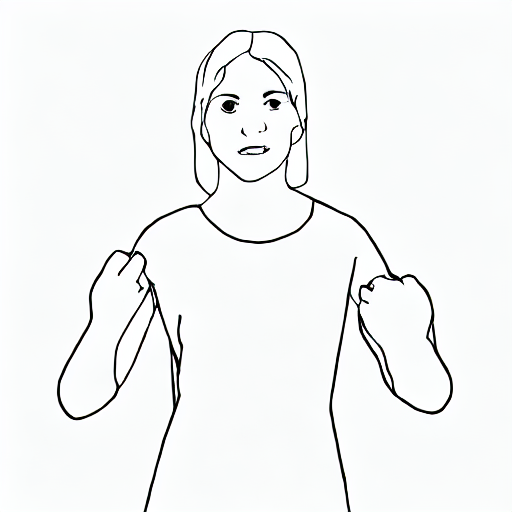}
    \end{minipage}
    \begin{minipage}{0.13\textwidth}
        \centering
        \includegraphics[width=\linewidth]{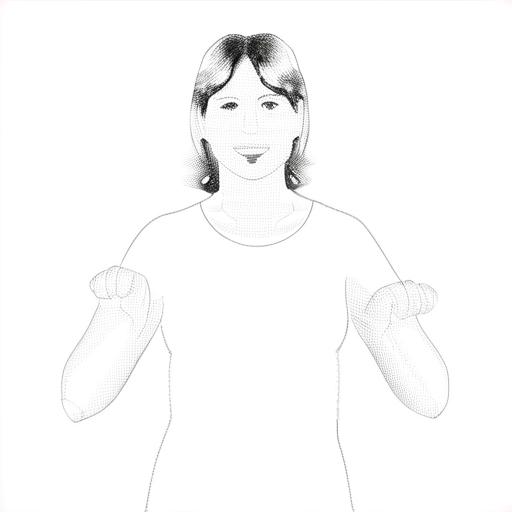}
    \end{minipage}

    \centering
    \begin{minipage}{0.13\textwidth}
        \centering
        \includegraphics[width=\linewidth]{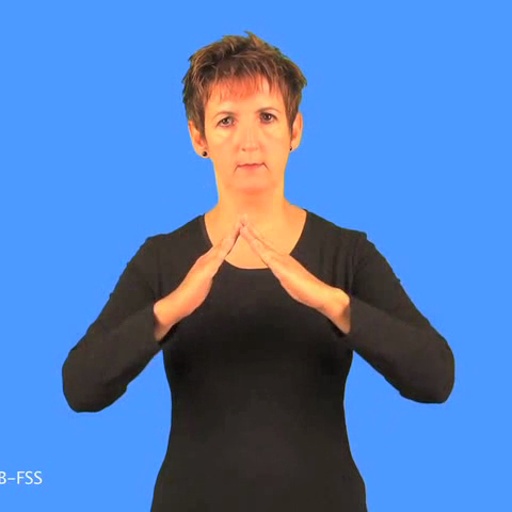}
    \end{minipage}
        \begin{minipage}{0.15\textwidth}
        \centering
        \includegraphics[width=\linewidth]{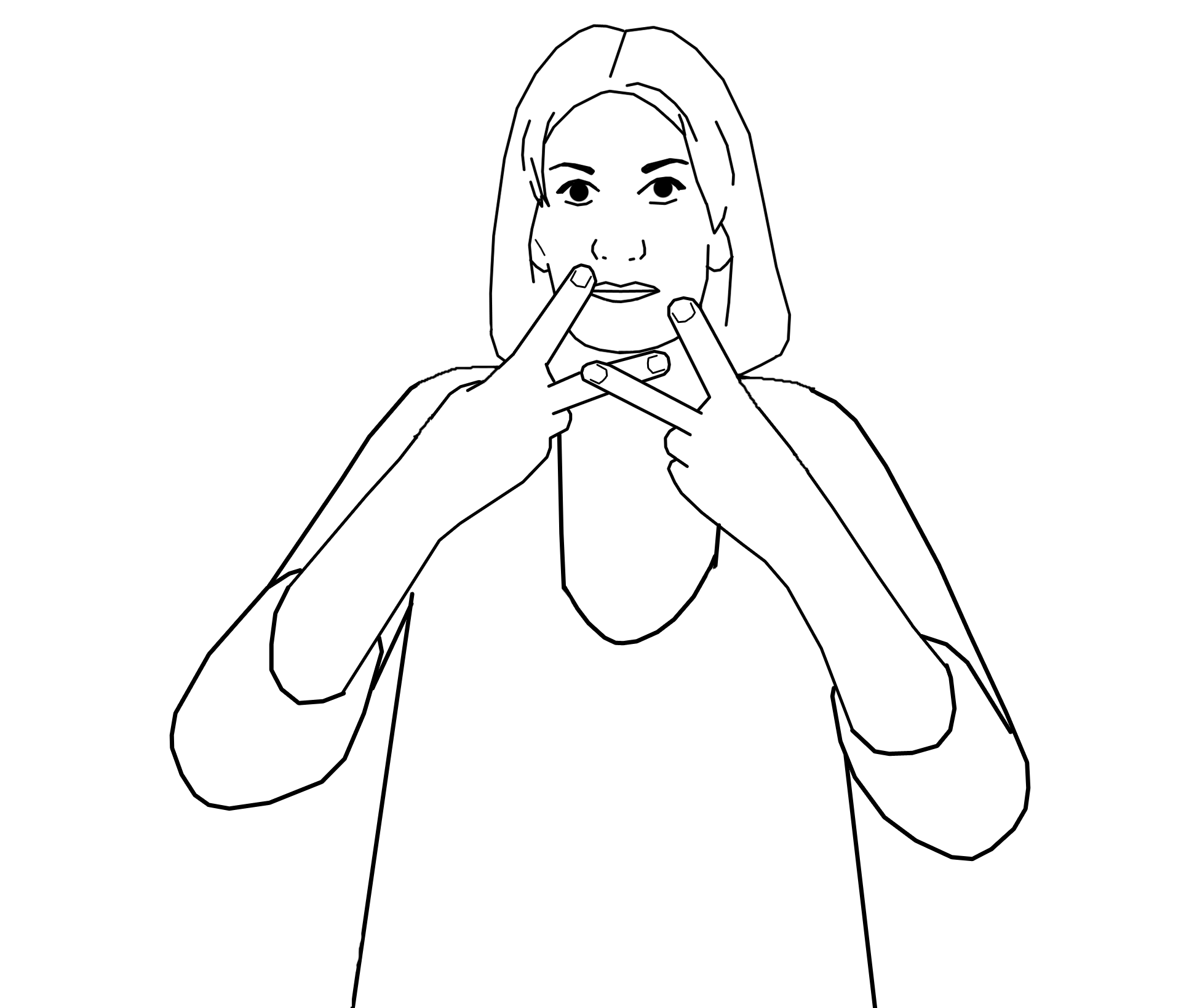}
    \end{minipage}
    \begin{minipage}{0.13\textwidth}
        \centering
        \includegraphics[width=\linewidth]{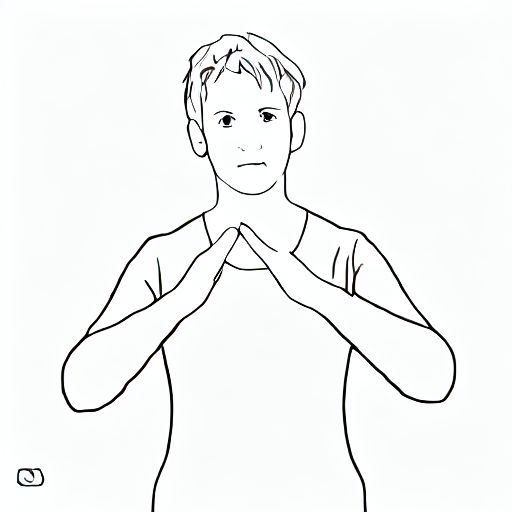}
    \end{minipage}
    \begin{minipage}{0.13\textwidth}
        \centering
        \includegraphics[width=\linewidth]{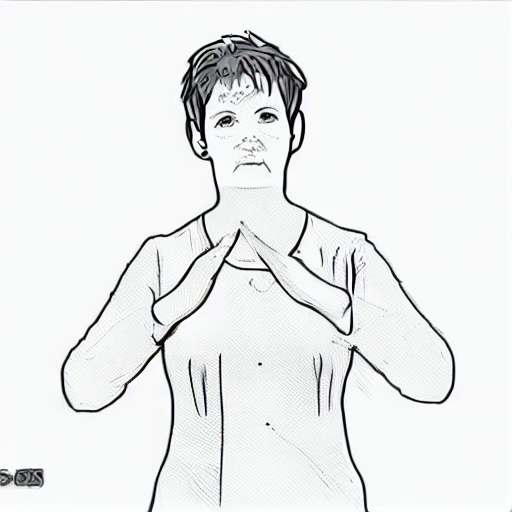}
    \end{minipage}
    \begin{minipage}{0.13\textwidth}
        \centering
        \includegraphics[width=\linewidth]{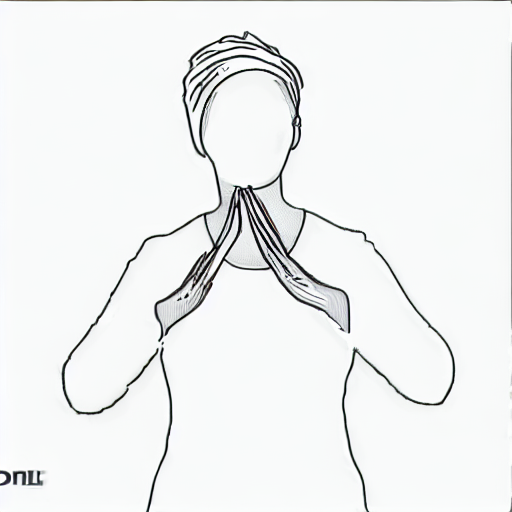}
    \end{minipage}
    \begin{minipage}{0.13\textwidth}
        \centering
        \includegraphics[width=\linewidth]{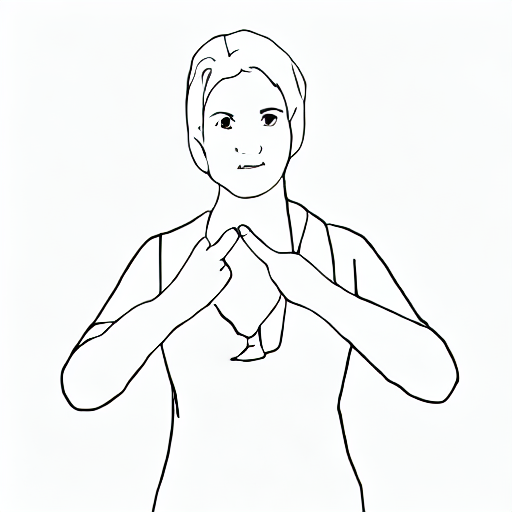}
    \end{minipage}
    \begin{minipage}{0.13\textwidth}
        \centering
        \includegraphics[width=\linewidth]{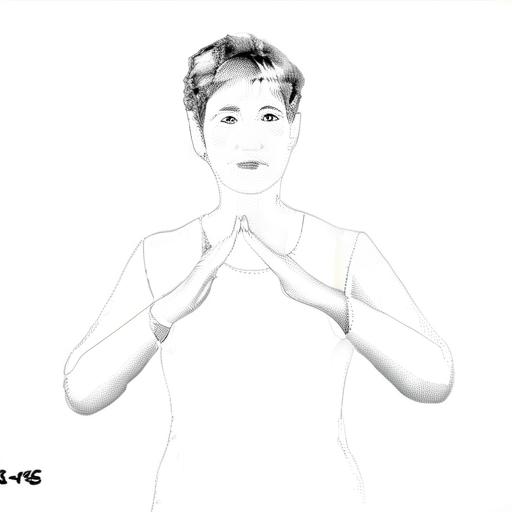}
    \end{minipage}

        \centering
    \begin{minipage}{0.13\textwidth}
        \centering
        \includegraphics[width=\linewidth]{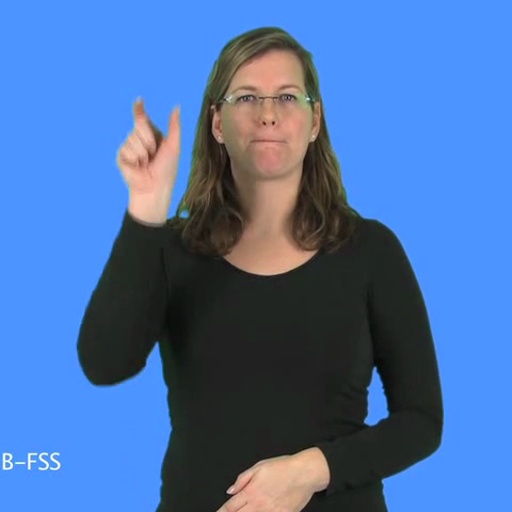}
    \end{minipage}
        \begin{minipage}{0.15\textwidth}
        \centering
        \includegraphics[width=\linewidth]{figures/style_images/32_no_arrow.png}
    \end{minipage}
    \begin{minipage}{0.13\textwidth}
        \centering
        \includegraphics[width=\linewidth]{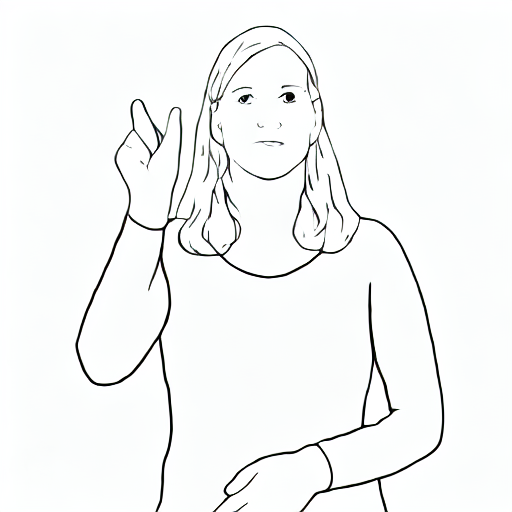}
    \end{minipage}
    \begin{minipage}{0.13\textwidth}
        \centering
        \includegraphics[width=\linewidth]{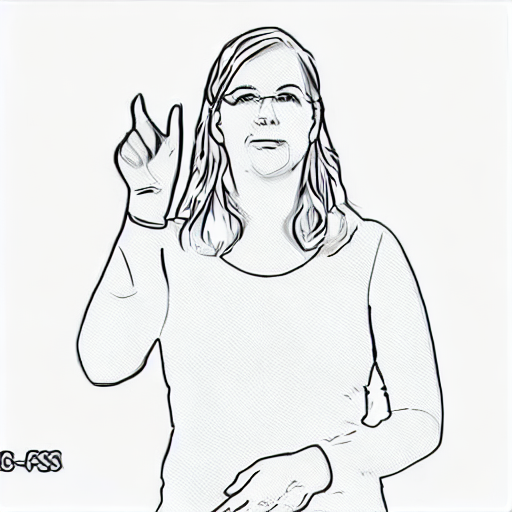}
    \end{minipage}
    \begin{minipage}{0.13\textwidth}
        \centering
        \includegraphics[width=\linewidth]{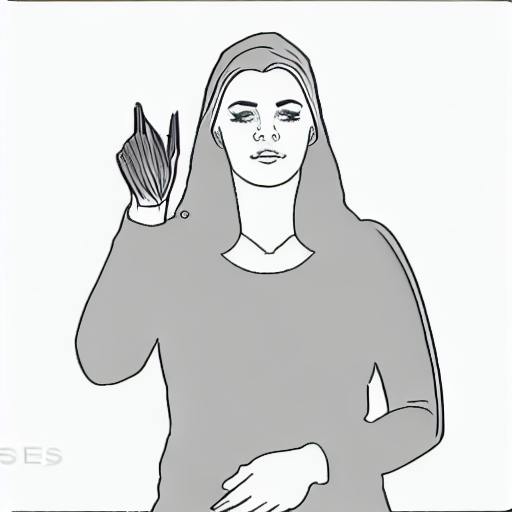}
    \end{minipage}
    \begin{minipage}{0.13\textwidth}
        \centering
        \includegraphics[width=\linewidth]{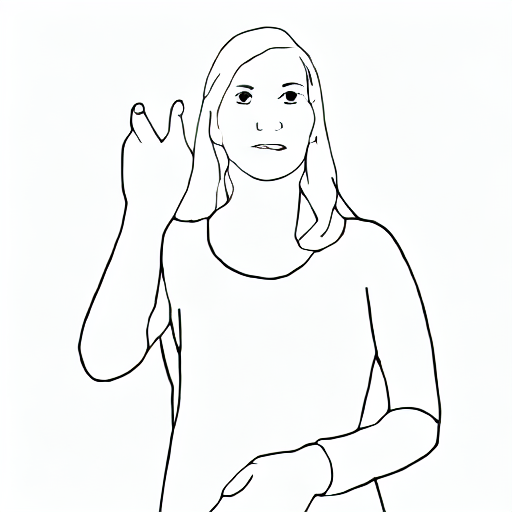}
    \end{minipage}
    \begin{minipage}{0.13\textwidth}
        \centering
        \includegraphics[width=\linewidth]{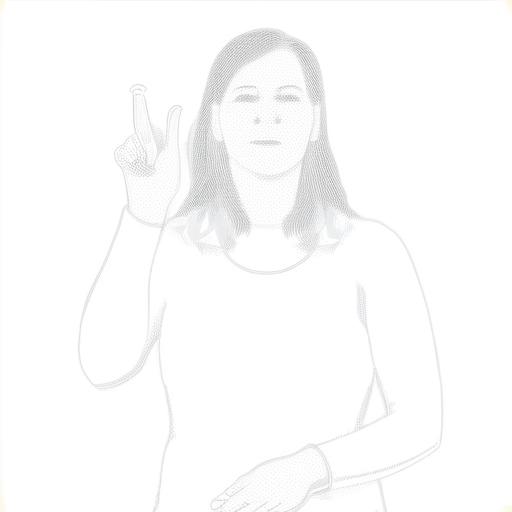}
    \end{minipage}

            \centering
    \begin{minipage}{0.13\textwidth}
        \centering
        \includegraphics[width=\linewidth]{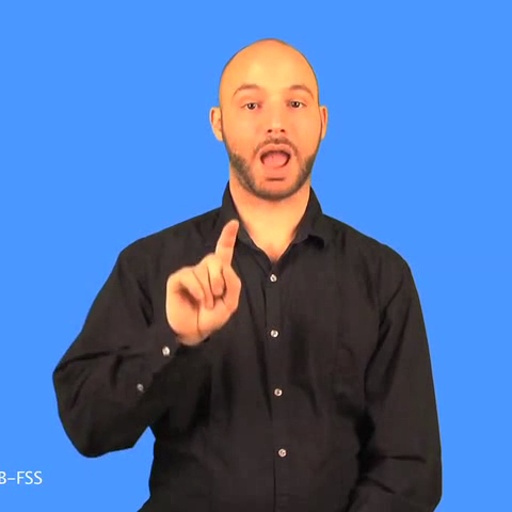}
        \par \text{Driving Image}
    \end{minipage}
        \begin{minipage}{0.15\textwidth}
        \centering
        \includegraphics[width=\linewidth]{figures/style_images/32_no_arrow.png}
        
        \par \text{Style Image}
        
    \end{minipage}
    \begin{minipage}{0.13\textwidth}
        \centering
        \includegraphics[width=\linewidth]{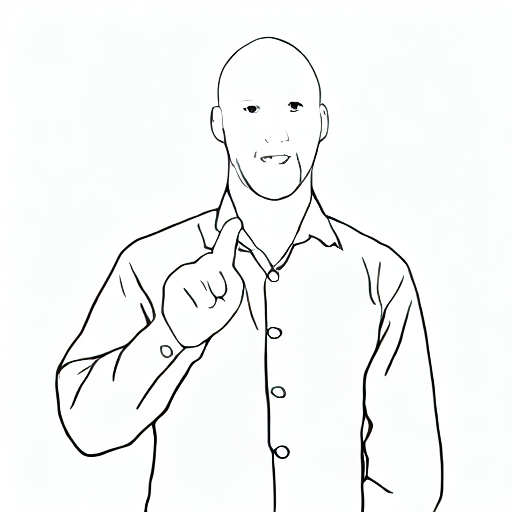}
        \par \text{Ours}
    \end{minipage}
    \begin{minipage}{0.13\textwidth}
        \centering
        \includegraphics[width=\linewidth]{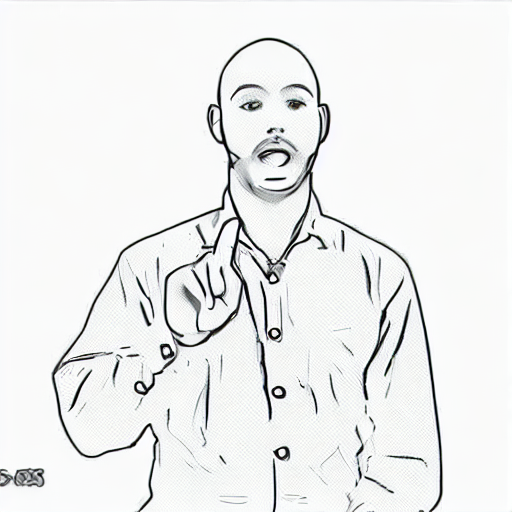}
        \par \text{StyleShot{\textsubscript{lineart}}}
    \end{minipage}
    \begin{minipage}{0.13\textwidth}
        \centering
        \includegraphics[width=\linewidth]{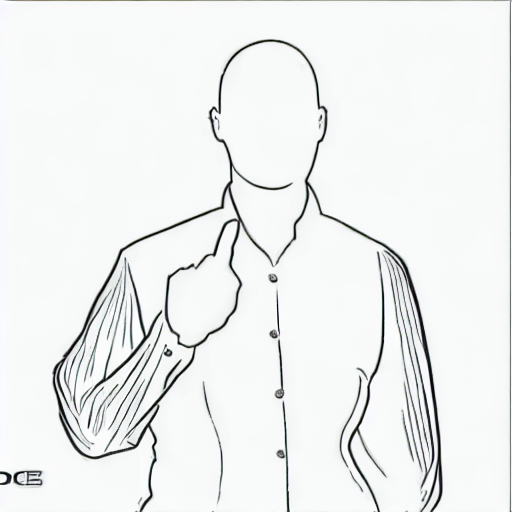}
        \par \text{StyleShot{\textsubscript{contour}}}
    \end{minipage}
    \begin{minipage}{0.13\textwidth}
        \centering
        \includegraphics[width=\linewidth]{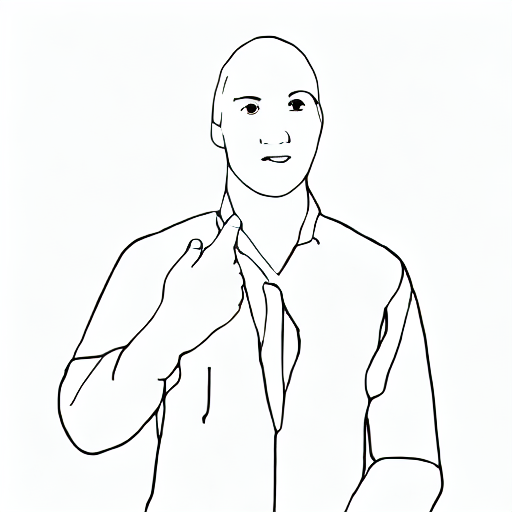}
        \par \text{Cross-Img}
    \end{minipage}
    \begin{minipage}{0.13\textwidth}
        \centering
        \includegraphics[width=\linewidth]{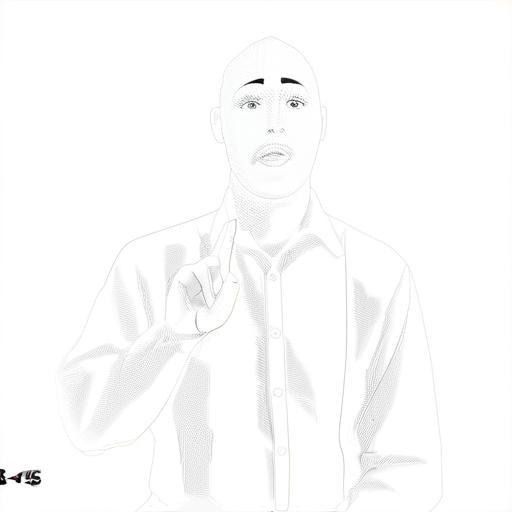}
        \par \text{InST}
    \end{minipage}

    \caption{Qualitative comparison of random samples from our data. Each input image is taken from a video frame as the driving image, and the style image corresponds to the illustration style. We compare our results to five baselines. In our method, we use two inputs: the input image (\textit{I\textsubscript{img}}) and an edge map (\textit{I\textsubscript{edges}}).}
    \label{fig:qualitative}

\end{figure*}

%% file: figures/different_styles.tex
\begin{figure*}[tb] %
    \centering
    \includegraphics[width=\textwidth]{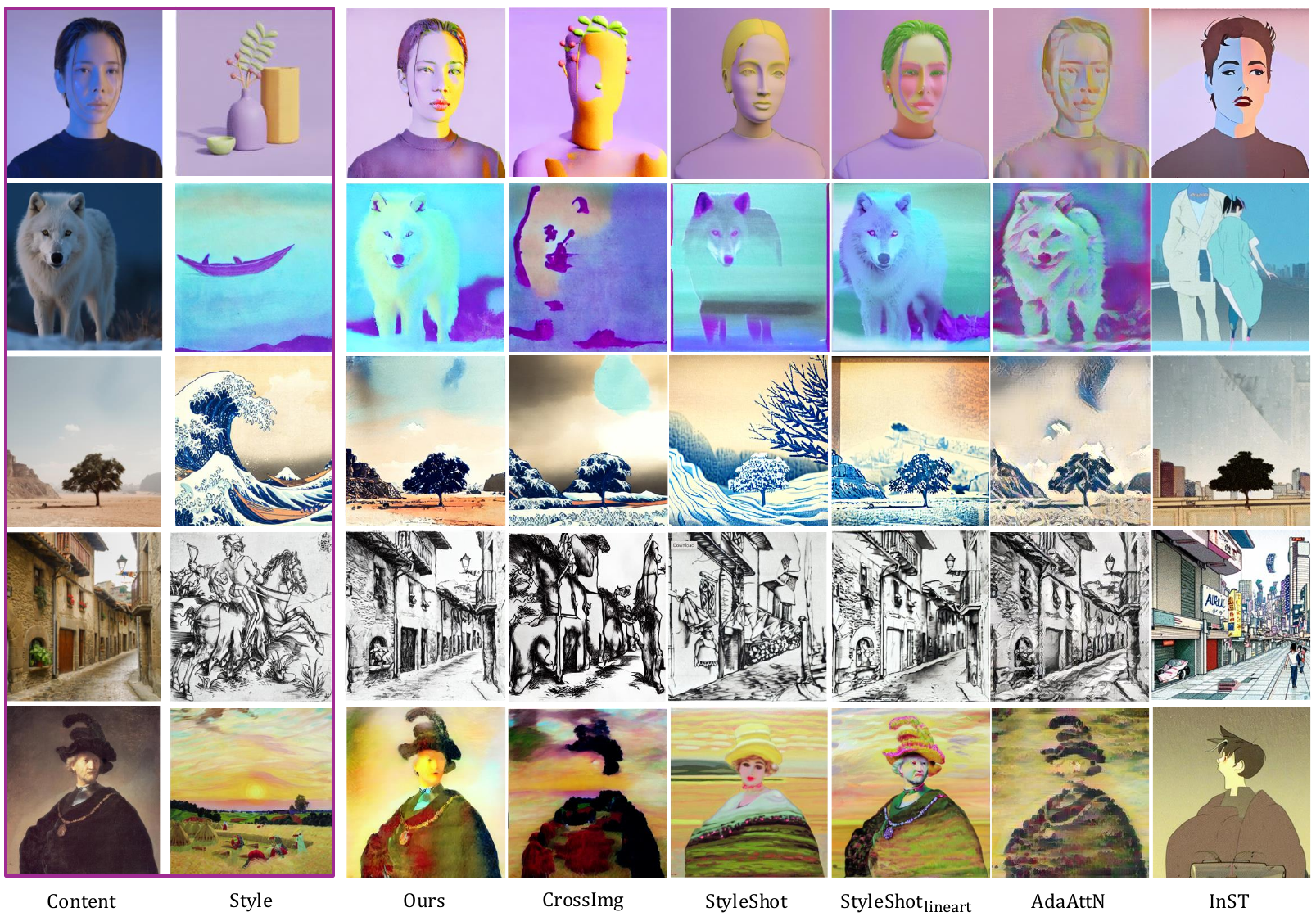}
        
    \caption{Generalization to Other Styles: We demonstrate that our method is comparable to state-of-the-art style transfer techniques, effectively generalizing across different domains and styles. Notably, it excels in preserving structural and fine details, particularly on subjects such as human faces.}
    \label{other_styles}
\end{figure*}

%% file: figures/ablations_queries.tex
\begin{figure*}[htb]
    \includegraphics[width=\linewidth]{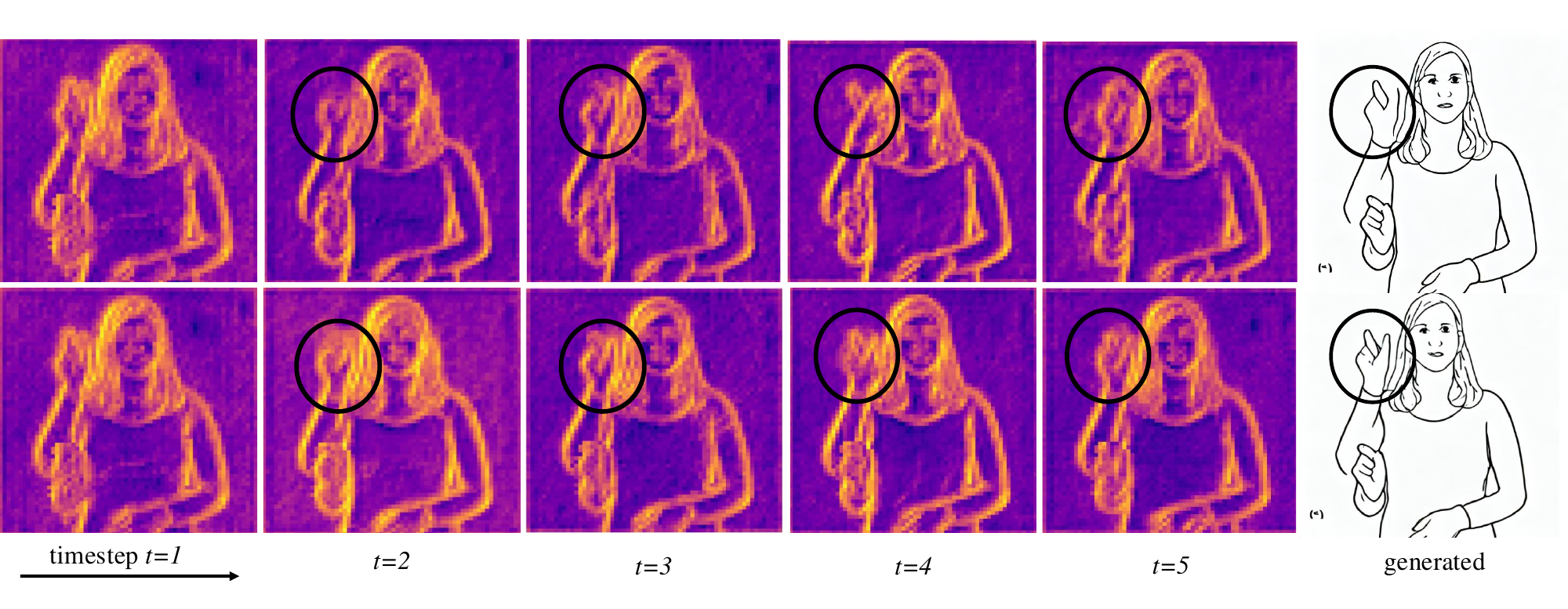}
    
    \caption{We visualize attention scores for a specific token during the early diffusion timesteps, highlighting the differences based on query initialization. In the top row, the overlay queries are derived from the current generated latent codes, while in the bottom row, the queries are initialized using the inverted \(Q_1\) and \(Q_2\) images.}
    \label{fig:queries_init}
\end{figure*}

%% file: tables/evaluation_metrics_style.tex
\newcolumntype{C}[1]{>{\centering\arraybackslash}p{#1}}

\begin{table}[tb]
\caption{\textbf{Style and Similarty Evaluation Metrics.} 
The best and second-best values are marked in \textbf{bold} and \underline{underlined}, respectively.}
\centering
\begin{tabular}{lccc}
\toprule
Method & {\small LPIPS $\downarrow$} & {\small Gram $\downarrow$} & {\small CLIPScore $\uparrow$} \\
\midrule
Our method & \underline{0.24} & \underline{3.58} & \textbf{0.87}  \\
AdaAttn~\cite{alaluf2023crossimage} & 0.28 & 6.43 & 0.83  \\
Cross-Image~\cite{liu2021adaattn} & \textbf{0.23} & \textbf{3.44} & \underline{0.86} \\
StyleShot\_Lineart~\cite{wang2024instantstyle} & 0.26 & 5.10 & 0.85 \\
StyleShot\_Contour~\cite{gao2024styleshot} & 0.28 & 4.90 & 0.82 \\
InstanceStyle~\cite{gao2024styleshot} & 0.26 & 9.61 & 0.79  \\
\bottomrule
\end{tabular}

\label{style_metric}
\end{table}

%% file: figures/masks_overlay.tex
\begin{figure}[tb] %
    \centering
    \begin{subfigure}{0.49\columnwidth} %
        \includegraphics[width=\linewidth]{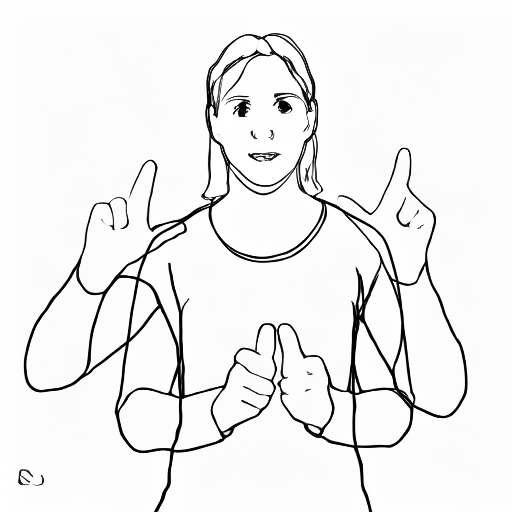}
        \caption{Pixelwise overlay}
    \end{subfigure}
    \begin{subfigure}{0.49\columnwidth}
        \includegraphics[width=\linewidth]{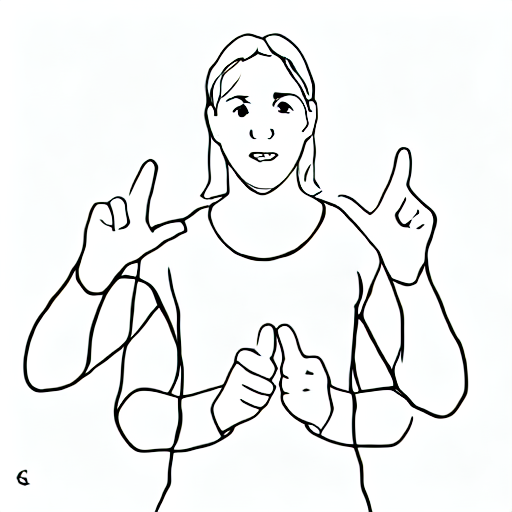}
        \caption{Ours final}
        
    \end{subfigure}
    \begin{subfigure}{0.49\columnwidth}
        \includegraphics[width=\linewidth]{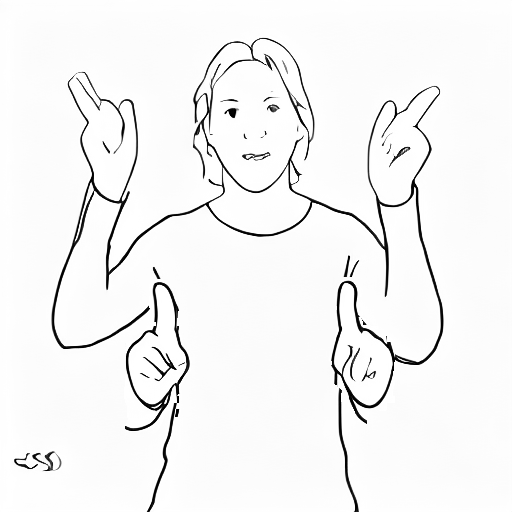}
        \caption{Pixelwise with Masks}
    \end{subfigure}
    \begin{subfigure}{0.49\columnwidth}
        \includegraphics[width=\linewidth]{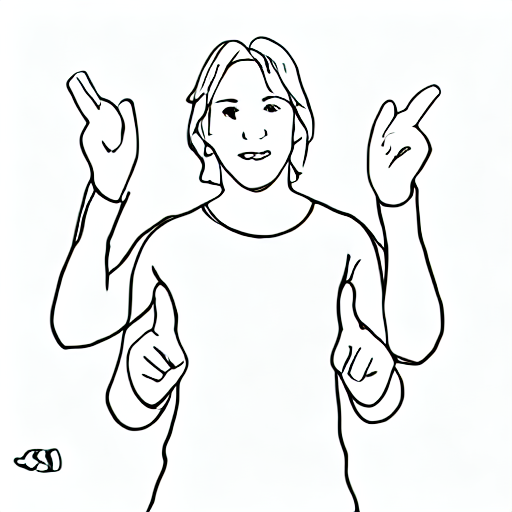}
        \caption{Ours final}
    \end{subfigure}
    
        \begin{subfigure}{0.49\columnwidth}
        \includegraphics[width=\linewidth]{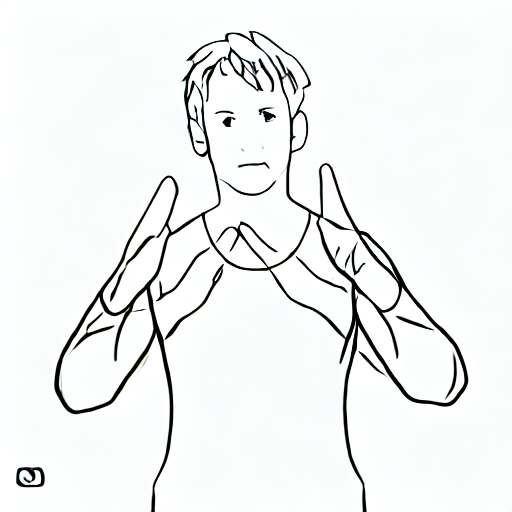}
        \caption{Ours without masks}
    \end{subfigure}
    \begin{subfigure}{0.49\columnwidth}
        \includegraphics[width=\linewidth]{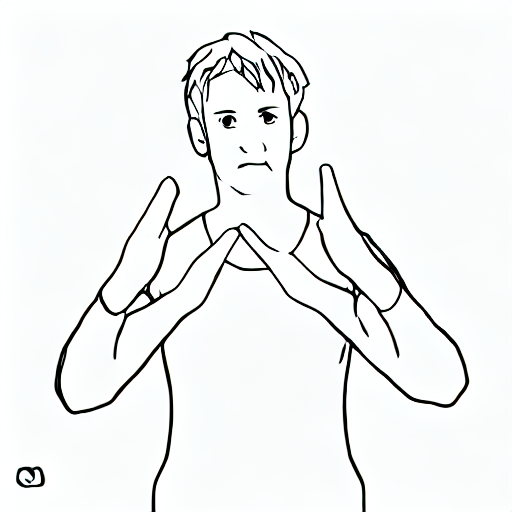}
        \caption{Ours final}
    \end{subfigure}

    \caption{We compare several approaches for creating the final overlay illustration. In (a), a simple pixel-space overlay results in double lines for the eyes and eyebrows. In (b), we show our final approach. In (c), we merge hands from the second image in pixel space, which can sometimes result in floating hands due to reliance on the body part segmentation model [19]. In (d), we show our final approach again. In (e), our overlay mechanism is applied, but it occasionally results in unclear hands. In (f), we improve the technique by adding hand masks.}
    \label{fig:masks}
\end{figure}

%% file: figures/ablation_inputs.tex
\begin{figure}[tb]
    \centering
    \begin{subfigure}{0.18\columnwidth} 
        \includegraphics[width=\linewidth]{{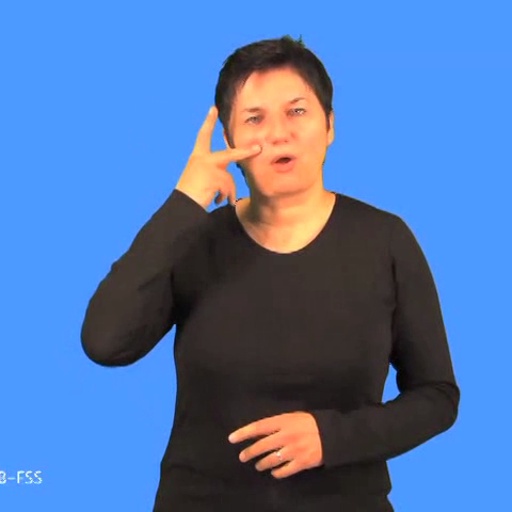}}
        \centering
        \par \tiny \text{Image} 
    \end{subfigure}
    \begin{subfigure}{0.18\columnwidth}
        \includegraphics[width=\linewidth]{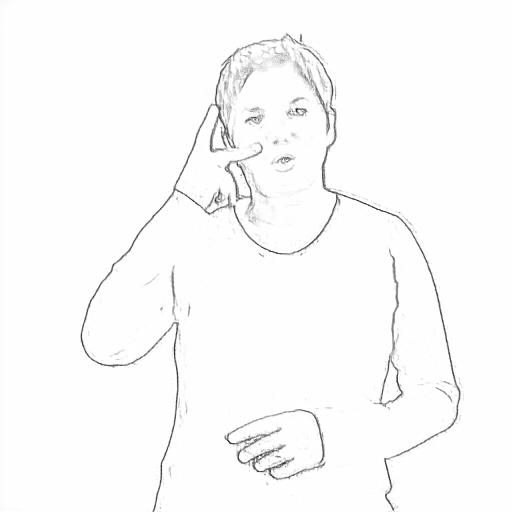}
        \par \tiny \text{Edges} 
        \centering
    \end{subfigure}
    \begin{subfigure}{0.18\columnwidth}
        \includegraphics[width=\linewidth]{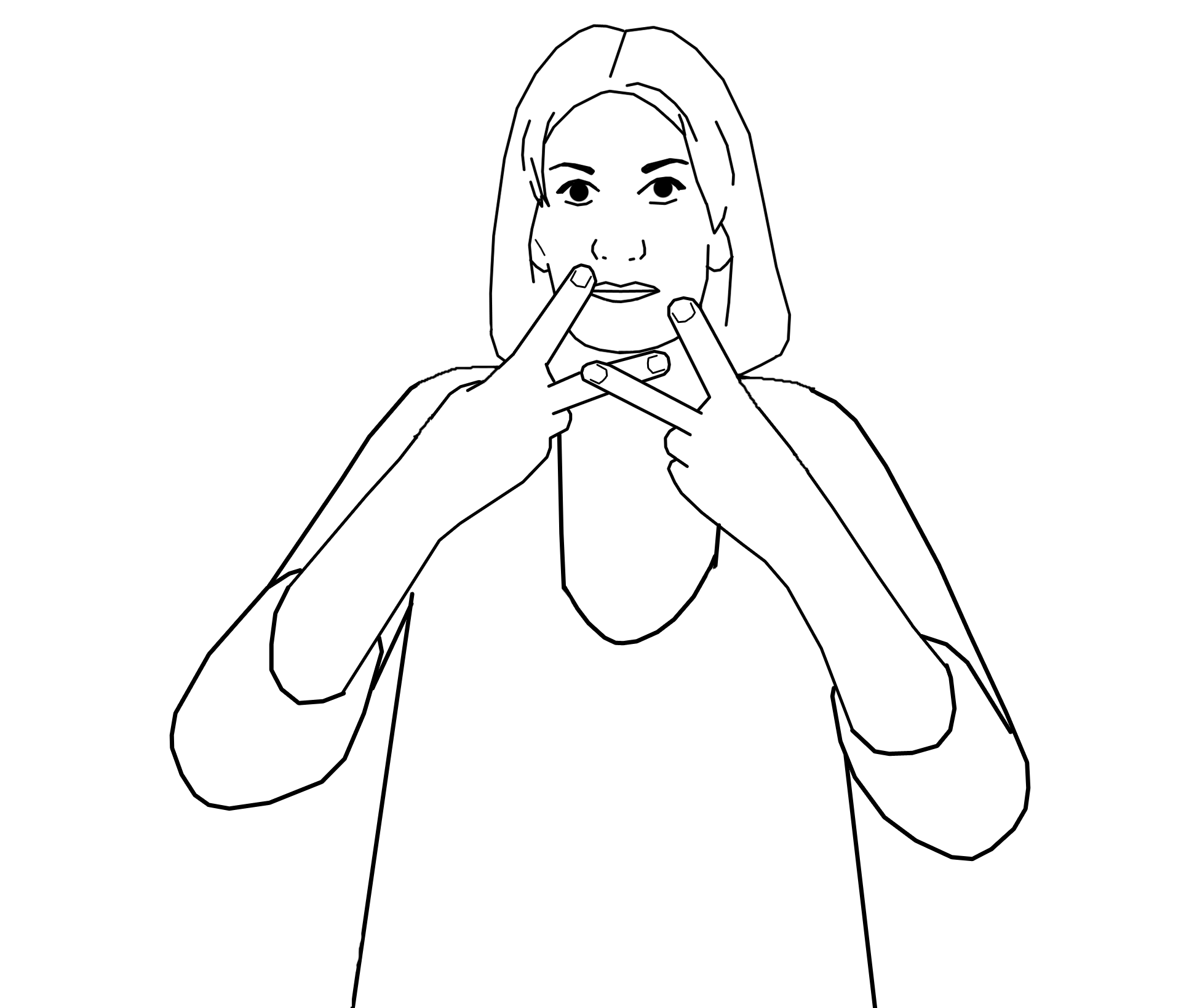}
        \centering
        \par \tiny \text{Style} 
    \end{subfigure}
    \begin{subfigure}{0.18\columnwidth}
        \includegraphics[width=\linewidth]{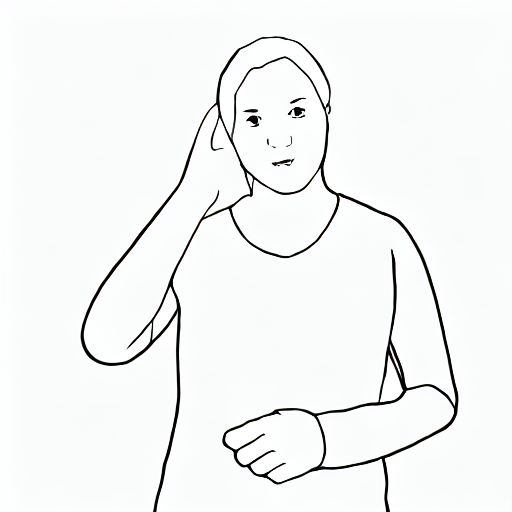}
        \centering
        \par \tiny \text{with $Q_{img}$} 
    \end{subfigure}
    \begin{subfigure}{0.18\columnwidth}
        \includegraphics[width=\linewidth]{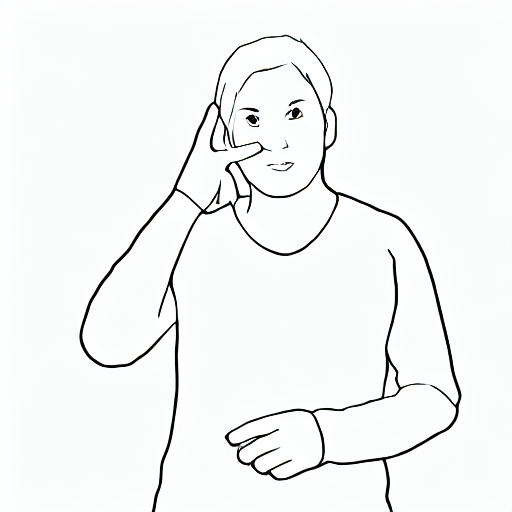}
        \par \tiny \text{with $Q_{edges}$} 
    \end{subfigure}
    \begin{subfigure}{0.18\columnwidth} 
        \includegraphics[width=\linewidth]{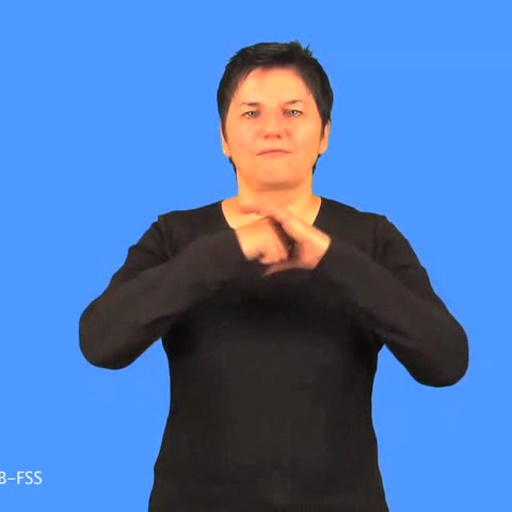}
        \centering
        \par \tiny \text{Image} 
    \end{subfigure}
    \begin{subfigure}{0.18\columnwidth}
        \includegraphics[width=\linewidth]{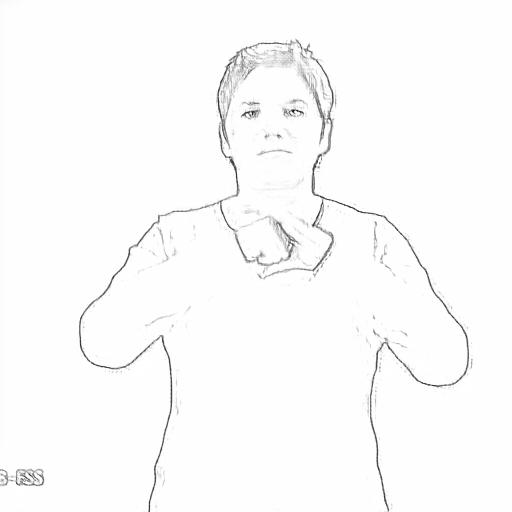}
        \centering
        \par \tiny \text{Edges} 
    \end{subfigure}
    \begin{subfigure}{0.18\columnwidth}
        \includegraphics[width=\linewidth]{figures/ablations/input_comparison/32_no_arrow.png}
        \par \tiny \text{Style} 
        \centering
    \end{subfigure}
    \begin{subfigure}{0.18\columnwidth}
        \includegraphics[width=\linewidth]{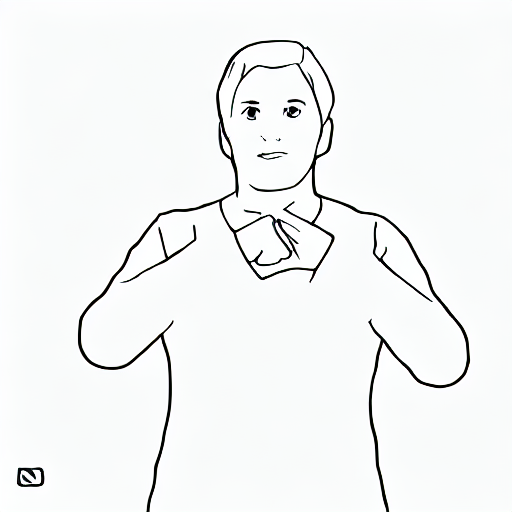}
        \centering
        \par \tiny \text{with $Q_{edges}$} 
    \end{subfigure}
    \begin{subfigure}{0.18\columnwidth}
        \includegraphics[width=\linewidth]{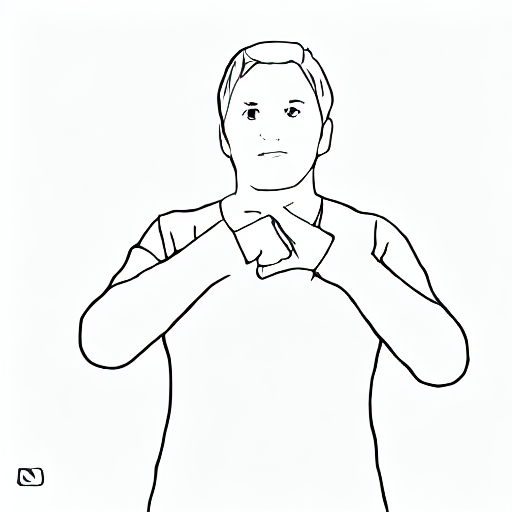}
        \centering
        \par \tiny \text{$Q_{img}, Q_{edges}$} 
    \end{subfigure}
    \caption{Different Choices of Query Injection: Injecting query edges improves hand accuracy but may lose details due to missing edge information, as seen in the top row; however, it may also lead to neglecting other details. Combining queries from both edges and the image helps recover lost details, such as the sleeves in the bottom-right image.}
    \label{fig:ablation_input}
\end{figure}